\documentclass[journal]{IEEEtran}
\pdfoutput=1 

\usepackage{graphics} %
\usepackage{epsfig} %
\usepackage{mathptmx} %
\usepackage{times} %
\usepackage{amsmath} %
\usepackage{amssymb}  %
\usepackage{diagbox} 
\usepackage{caption}
\usepackage{subfigure}
\usepackage{graphicx}
\usepackage{hyperref}
\usepackage{bm}
\usepackage{color}
\usepackage{float}

\ifCLASSINFOpdf
\else
\fi

\begin{document}
\title{{Flying Through a Narrow Gap Using End-to-end Deep Reinforcement Learning Augmented with Curriculum Learning and Sim2Real}}

\author{Chenxi Xiao$^{1}$$^{\dagger}$, Peng Lu$^{2*}$$^{\dagger}$ and Qizhi He$^{3}$ 
\thanks{* Corresponding author}%
\thanks{$\dagger$ These authors contributed equally to this manuscript.}%
\thanks{$^{1}$ Chenxi Xiao was with the Hong Kong Polytechnic University and is currently with Purdue University, 
        {\tt\small xiao237@purdue.edu}}%
\thanks{$^{2}$ Peng Lu is currently with the Adaptive Robotic Controls Lab at the University of Hong Kong and was previously with the Hong Kong Polytechnic University.
        {\tt\small lupeng@hku.hk}}%
\thanks{$^{3}$ Qizhi He is with Northwestern Polytechnical University, 
        {\tt\small  heqizhi@mail.nwpu.edu.cn}}%
}

\markboth{IEEE TRANSACTIONS ON NEURAL NETWORKS AND LEARNING SYSTEMS}
{Shell \MakeLowercase{\textit{et al.}}: Bare Demo of IEEEtran.cls for IEEE Journals}

\maketitle

\begin{abstract}
Traversing through a tilted narrow gap is previously an intractable task for reinforcement learning mainly due to two challenges. First, searching feasible trajectories is not trivial because the goal behind the gap is difficult to reach. Second, the error tolerance after Sim2Real is low due to the relatively high speed in comparison to the  gap's narrow dimensions. This problem is aggravated by the intractability of collecting real-world data due to the risk of collision damage. In this paper, we propose an end-to-end reinforcement learning framework that solves this task successfully by addressing both problems. To search for dynamically feasible flight trajectories, we use a curriculum learning to guide the agent towards the sparse reward behind the obstacle.  To tackle the Sim2Real problem, we propose a Sim2Real framework that can transfer control commands to a real quadrotor without using real flight data. To the best of our knowledge, our paper is the first work that accomplishes successful gap traversing task purely using deep reinforcement learning.
\end{abstract}


\IEEEpeerreviewmaketitle

\section{Introduction} 
\subsection{Problem Background} \label{intro}

Aggressive flight can enhance the maneuverability of {quadrotors}. For instance, in search and rescue applications, quadrotors are required to explore {unstructured environments} with narrow entries. The {quadrotor's} activity range can be enlarged if it is capable of flying through narrow gaps that are considered {intractable} {from the aspect of} a classical flight controller. Moreover, consider that quadrotors powered by batteries have a limited activity range. {Avoiding obstacles by taking shortcuts through narrow gaps may {reduce the power consumption} by avoiding taking long detours.} Accordingly, it is necessary to develop motion planners that can perform aggressive motions.

{However, constrained flight motions can be difficult to perform due to the quadrotor's underactuated dynamics}. One example is to fly through a tilted rectangular gap, during which the quadrotor needs to avoid collision and {subject to} attitude and position constraints simultaneously. However, keeping a tilted attitude may induce a large horizontal acceleration. {As a result,} it would generate a horizontal position shift and increase the chance of colliding with the bezel. Therefore, finding a feasible trajectory is not trivial.

Aggressive flight planning has been explored for decades \cite{3,4}. {Conventional studies} model the problem as a constrained motion planning problem that can be solved by optimizing manually defined loss functions. However, all these approaches have to simplify the problem using strong mathematical assumptions {so that it can be formulated under the optimal control paradigm}. During optimization, excessive prior knowledge is added (refer to Sec.~\ref{rw:gap}), {such} that motions that are inconsistent with the priors will be penalized during optimization. Accordingly, only solutions of a few specific patterns can be 
obtained, which eliminated the possibility of {obtaining} a solution of better patterns. %

{
Compared to previous works, model-free reinforcement learning mainly has two advantages. {Firstly, control policy can be optimized directly using unstructured environments and also under the quadrotor's strong non-linear dynamics.} {Convexity of loss function is still desired but is no longer a prerequisite}. Secondly, the solution pattern is not biased by the aforementioned handcrafted priors. Instead, the model-free learning paradigm only relies on a reward function that indicates whether the goal has been reached. Therefore, leveraging on the reinforcement learning makes it possible to get a solution that is better than the sub-optimal trajectory from the user-defined {solution} space.
}

\begin{figure}[t]
        \centering
	\includegraphics[scale=0.48]{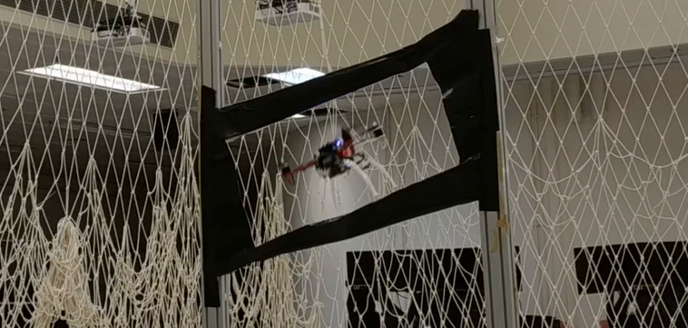}
        \caption{The quadrotor controlled by reinforcement learning is passing through a tilted narrow gap. }
        \label{concept}
        \vspace{-6mm}
\end{figure}

\subsection{Contributions} \label{contribution}

In this paper, {we propose an end-to-end reinforcement learning solver for the quadrotor's gap traversing task. Our approach does not rely on excessive problem-orientated priors.} The methodological contributions {are mainly from} two aspects. First, due to the limited exploration ability of current reinforcement learning algorithms, searching for a feasible trajectory is not trivial. To this end, we {propose to guide the exploration using curriculum learning, with which we have acquired feasible trajectories without conventional model-based motion planners}. Second, transferring our learned policy to a real quadrotor is challenging due to the {low error tolerance of Sim2Real as well as the intractability of {collecting} real trajectories. To tackle this issue, We propose a novel Sim2Real approach that enables successful Sim2Real transfer without using real flight trajectories.}

{
\section{Related Works}
\subsection{Drone control and planning by reinforcement learning}
Reinforcement learning is reportedly a powerful approach for various flight control and planning tasks. Zhang et al. \cite{5} applied Guided Policy Search (GPS) to a quadrotor collision avoidance task. The policy from GPS can outperform offline iterative Linear–Quadratic–Gaussian (LQG) planner and Model Predictive Control (MPC) planner  {using an ideal quadrotor model, and a model with 5 percentage mass error.} Hwangbo et al. \cite{6} demonstrated a method to train a reinforcement learning policy that can control a real-world quadrotor from the motor thrust level. The learned policy can accomplish hovering, waypoint tracking and posture stabilization from random initial states. Molchanov et al. \cite{16} stabilized a quadrotor using Proximal Policy Optimization (PPO) algorithm. Lambert et al. \cite{17} stabilized a quadrotor based on a model-based reinforcement learning approach. Li et al. \cite{9} designed a reinforcement learning policy that can track targets. Mannucci et al. \cite{Mannucci2018} proposed two reinforcement learning-based algorithms to control the attitude of the aircraft.

\subsection{Quadrotor traversing through a narrow gap} \label{rw:gap}
{In previous studies, the gap-traversing task has been solved by Falanga et al.\cite{3} and Loianno et al.\cite{4} based on the conventional optimal control framework }. To be specific, Falanga et al.\cite{3} modeled the problem as an optimization problem on a pre-defined trajectory set. The trajectory is constrained to be a quadratic function that must intersect with the gap center. The vehicle's velocity and acceleration during traversing are manually defined, and the time duration of traversing is to be reduced by optimization. However, the excessive hand-crafted prior knowledge may stifle better solutions to be obtained, since there is no evidence that the involved constraints are able to generate optimal solutions. The method is also problem-orientated and not scalable to {unstructured} environments with additional obstacles or irregular {gap dimensions}. 
Similarly, Loianno et al.\cite{4} also defined the problem in an optimal control paradigm with excessive priors i.e. a parabolic trajectory, constant motor thrust, zero angular velocity during traversing, and a fixed trajectory starting point. 

To {reduce} the dependencies on the aforementioned priors, literature \cite{20} {is the first known study that implemented gap traversing using reinforcement learning. A neural network is utilized} to imitate trajectories from an optimal control solver. The solution was then fine-tuned by training in AirSim \cite{airsim}. The final trajectory pattern is reportedly more diverse than the parabolic curve trajectories from previous studies \cite{3, 4}. However, the {initial trajectory being cloned is still obtained from the optimal control framework with excessive priors. It is known that imitation learning may still end up with local optimal solutions that are similar to demonstrations without sufficient exploration \cite{exploration-imit}.  Besides, the method is still not detached from optimal control that requires excessive priors.  Therefore, a pure reinforcement learning solver that can solve the problem in an end-to-end paradigm is desired. To the best of our knowledge, our work is the first instance of work that only uses a model-free reinforcement learning solver to accomplish this gap-traversing task in the real world. }
}

\section{Task and Method Overview}
\subsection{Task Statement}
{
Our task is to plan aggressive trajectories for passing through a tilted narrow hole, as demonstrated in Fig.~\ref{concept}.
}

\begin{figure}[h]
\centering
\subfigure[]{\includegraphics[width=4.2cm, height=3.5cm, angle=0]{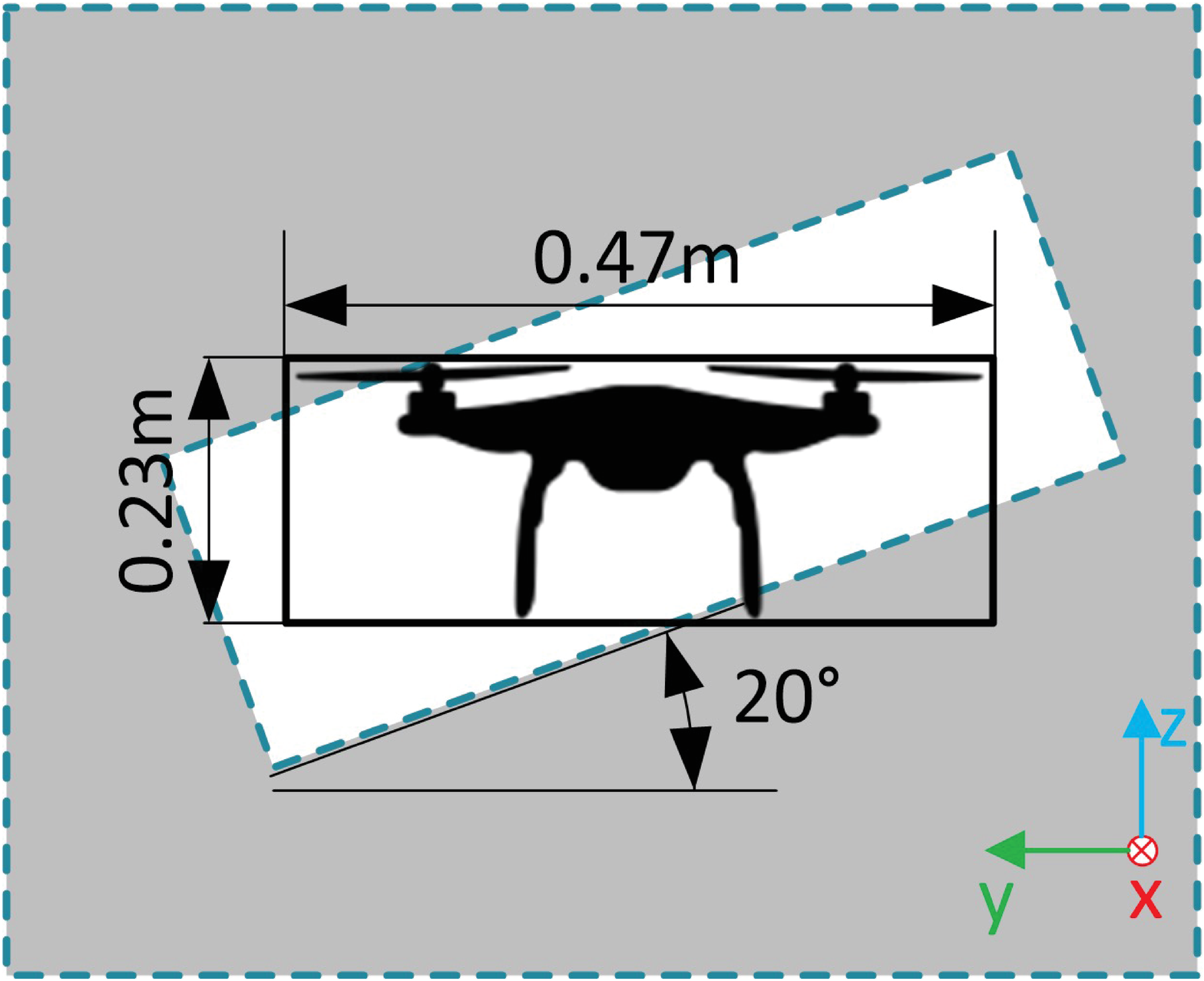}}
\subfigure[]{\includegraphics[width=4.2cm, height=3.5cm, angle=0]{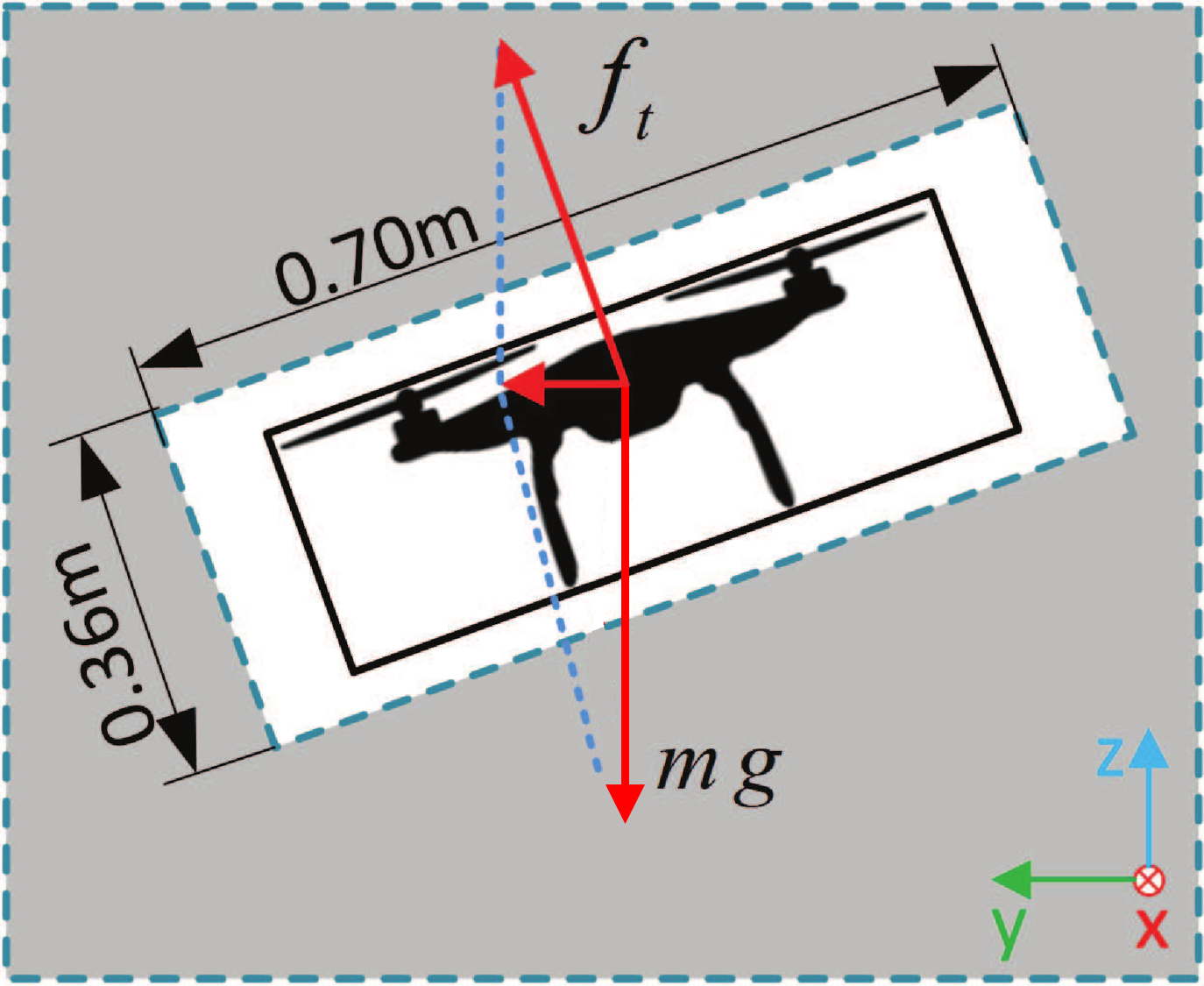}}
\caption{(a) A direct traverse that cannot be accomplished. (b) One possible traverse which avoids collision with the gap. However, the horizontal acceleration may lead to collision.} 
\label{hole_geo}
\end{figure}

A direct traverse is not feasible, as shown in Fig.~\ref{hole_geo} (a). The black rectangle is the bounding box of the quadrotor. The gray background rectangle represents a wall with a tilted gap. Fig.~\ref{hole_geo} (b) shows an instance in which the geometric constraint is satisfied. But the joint force induced by {motor thrusts and quadrotor's gravity} will lead to additional horizontal acceleration that may lead to a collision, as shown by red arrows. In addition, the pitch angle used for dashing forward will increase the lateral area of the quadrotor's bounding box, which reduces the safe distance margin.

\begin{figure}[htb]
\vspace{-4mm}
\centering
\includegraphics[scale=0.35]{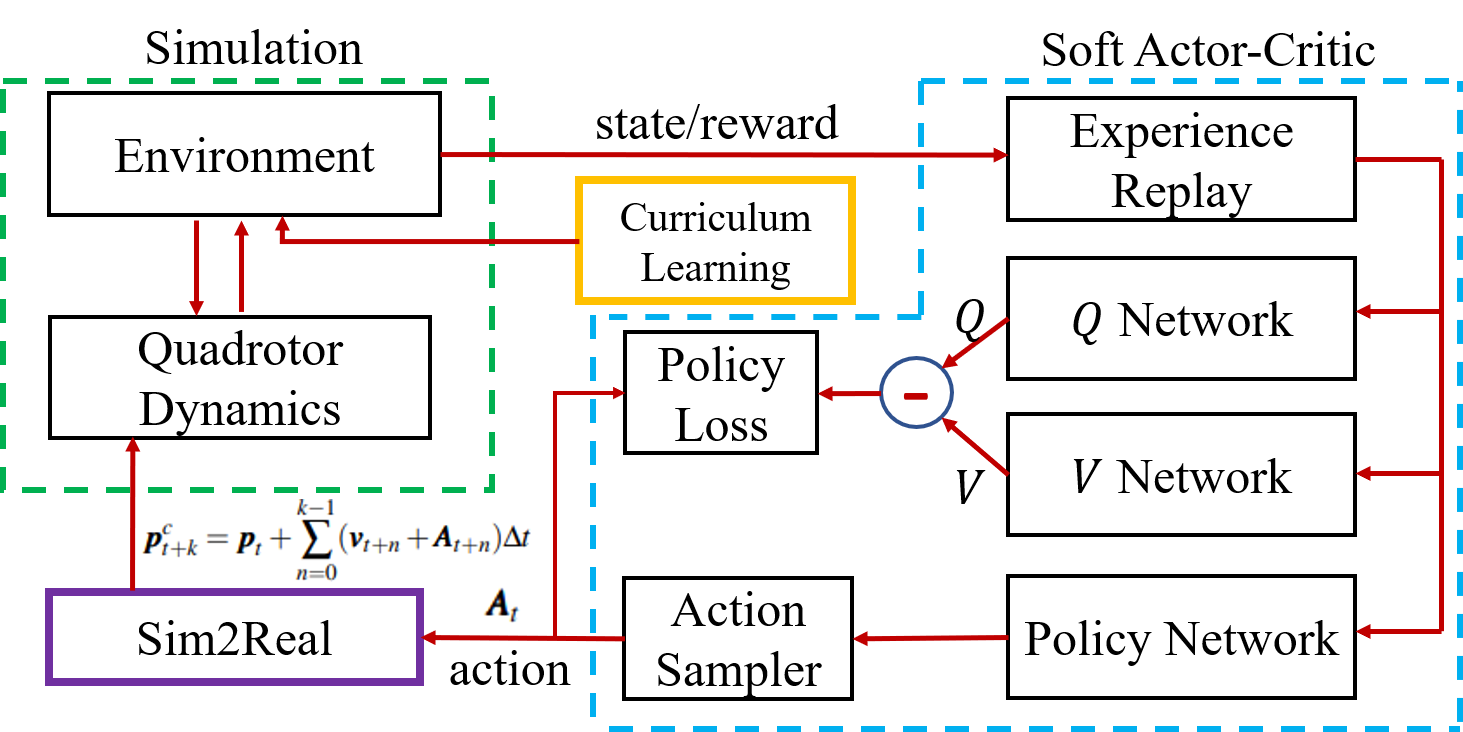}
\caption{An overview of our proposed training framework. }
\label{software_overview}
\vspace{-4mm}
\end{figure}

 {We demonstrate our training framework in Fig.~\ref{software_overview}}. These modules (Simulation, Soft Actor-Critic, Sim2Real) are discussed in Section IV, V, VI, respectively. 
 
\section{Simulation Environment for Reinforcement Learning}

One shortcoming of model-free reinforcement learning is the low data efficiency. {Training the policy directly in real world is impractical because a real quadrotor is too fragile to endure a large number of failure rollouts. Instead, a simulation environment is created using the dynamics described in Sec.~\ref{dynamics}. }

\subsection{Quadrotor Dynamics} \label{dynamics}
{We model the quadrotor as a rigid body with non-linear dynamics \cite{13}}. The angular acceleration is modeled as Eq.~(\ref{eq_omega}).
\begin{equation}
\left[\begin{array}{c}{\dot{\omega_x}} \\ {\dot{\omega_y}} \\ {\dot{\omega_z}}\end{array}\right]=\left[\begin{array}{c}{\tau_{\phi} I_{x x}^{-1}} \\ {\tau_{\theta} I_{y y}^{-1}} \\ {\tau_{\psi} I_{z z}^{-1}}\end{array}\right]-\left[\begin{array}{c}{\frac{I_{y y}-I_{z z}}{I_{x x}} \omega_y \omega_z} \\ {\frac{I_{z z}-I_{x x}}{I_{y y}} \omega_x \omega_z} \\ {\frac{I_{x x}-I_{y y}}{I_{z z}} \omega_x \omega_y}\end{array}\right]
\label{eq_omega}
\end{equation}
$\tau_{\phi}$, $\tau_{\theta}$, $\tau_{\psi}$ are the roll, pitch and yaw torques, respectively. $I_{x x}$, $I_{y y}$, $I_{z z}$ are the rotational inertia of x, y and z axis in the body frame. $\omega_x$, $\omega_y$, $\omega_z$ are the roll, pitch and yaw rates, respectively.

Similarly, we model the translational motion as Eq.~(\ref{linear}). $m$ is the quadrotor's mass, $\bm{a}$ is the rigid body linear acceleration, $g$ is the gravitational constant.  $f_{t}$ is the total thrust. $\bm{e}_3=[0,0,1]^T$, $\bm{R}$ is the rotational matrix from body to earth frame, $\bm{f_{d}}$ is the drag force induced by linear motions, {which is proportional to the squared body linear velocity $v_{x}$, $v_{y}$, $v_{z}$ in its x, y, z axis, respectively \cite{13b}.} 
\begin{equation}
m \bm{a}=m \bm{e}_3g+ \bm{R} \bm{e}_3 f_{t}+ \bm{f}_{d}
\label{linear}
\end{equation}

We use control distribution matrix to {model the mapping relationship} from motor {thrusts} to $\tau_{\phi}$, $\tau_{\theta}$, $\tau_{\psi}$ and $f_{t}$, as shown in Eq.~(\ref{mixer}). $C_{\mathrm{T}}$ is the thrust coefficient, and $C_{\mathrm{M}}$ is the torque coefficient. Note that the control distribution matrix corresponds to the X type quadrotor and therefore the arm length is $\frac{\sqrt{2}}{2}d$, where $d$ is the horizontal side length of the Oriented Bounding Box (OBB). 
\begin{equation}
\setlength{\arraycolsep}{1.25pt}
\renewcommand\arraystretch{1.5}
{\small
\left[\begin{array}{c}{f_{t}} \\ {\tau_{\phi}} \\ {\tau_{\theta}} \\ {\tau_{\psi}}\end{array}\right]=
\left[\begin{array}{cccc}{C_{\mathrm{T}}} & {C_{\mathrm{T}}} & {C_{\mathrm{T}}} & {C_{\mathrm{T}}} \\ {\frac{\sqrt{2}}{2} d C_{\mathrm{T}}} & {-\frac{\sqrt{2}}{2} d C_{\mathrm{T}}} & {-\frac{\sqrt{2}}{2} d C_{\mathrm{T}}} & {\frac{\sqrt{2}}{2} d C_{\mathrm{T}}} \\ {\frac{\sqrt{2}}{2} d C_{\mathrm{T}}} & {\frac{\sqrt{2}}{2} d C_{\mathrm{T}}} & {-\frac{\sqrt{2}}{2} d C_{\mathrm{T}}} & {-\frac{\sqrt{2}}{2} d C_{\mathrm{T}}} \\ {C_{\mathrm{M}}} & {-C_{\mathrm{M}}} & {C_{\mathrm{M}}} & {-C_{\mathrm{M}}}\end{array}\right]\left[\begin{array}{c}{\omega_{1}^{2}} \\ {\omega_{2}^{2}} \\ {\omega_{3}^{2}} \\ {\omega_{3}^{2}}\end{array}\right]
} 
\label{mixer}
\end{equation}

\subsection{Environmental State Variables}

The state variable $\bm{s}$ used in the reinforcement learning {constitutes of} the following information: linear position error towards the goal state ($p^e_x, p^e_y$ and $p^e_z$), linear velocities ($v_x, v_y$ and $v_z$), roll and pitch angles ($\phi$ and $\theta$), roll and pitch rates ($\omega_{x}$ and $\omega_{y}$). Note that we do not implement control on the yaw channel and therefore we do not feed yaw information to the network. Each entry of the linear position error vector $\bm{p}^e$ is defined as:
\begin{equation}
\renewcommand\arraystretch{1.5}
{{p}_{i}^{e}}=\operatorname{sign}\left(p_{i} - p_{G_i}\right) \sqrt{\left|p_{i} - p_{G_i}\right|}
\label{dis_err}
\end{equation}
Subscript $i$ corresponds to the $x$, $y$ and $z$ position channel. $\bm{p}$ is the robot position, and $\bm{p_{G}}$ is the position of the goal point (defined in the world frame, $\bm{p_{G}}$ is a fixed point located at 25 centimeters behind the gate's central point). Eq.~(\ref{dis_err}) magnifies the positional error when the quadrotor is close to the gate's center, {aiming to enhance the discriminability of the positional feedback {in that case}}.

\subsection{Reward Design}

We use a simple reward function because we do not intend to restrict the solution space by excessive prior knowledge (discussed in Sec.~\ref{intro}). We use a +1,000 value as the goal reward. This reward can only be acquired {if the quadrotor passes through the hole without any collisions detected}. The reward scaling is from parameter tuning. However, due to the difficulty of visiting the states behind the gate, this {goal} reward itself is too sparse to guide the training. An auxiliary penalty reward that is negative proportional to the distance $-\left\|\bm{p}-\bm{p}_{G}\right\|_2$ is also used. {This penalty reward encourages the quadrotor to move towards the target and therefore significantly improved the training stability. Note that this auxiliary reward accumulated in the whole episode is much smaller than the goal reward because the solution should not be dominant by this auxiliary reward. Overall, the reward function $r(\bm{p})$ is given in Eq.~(\ref{rr})}
\begin{equation}
r(\bm{p})=\left\{\begin{array}{cc}{+1,000} & {(\text {when goal is reached})} \\ {-\left\|\bm{p}-\bm{p}_{G}\right\|_2} & {\text {(otherwise) }}\end{array}\right. 
\label{rr}
\end{equation}

\subsection{Simulated Gap}

The environment includes a wall with a narrow gap. We terminate the simulation episode {immediately when a} collision between the quadrotor and the wall is detected. For this, we implemented a simple collision checker. The intersection points between the bounding box of the quadrotor and the wall are calculated {in real-time. One} collision is recognized if any intersection points are outside the gap's boundary. {A traversing attempt is successful if no collision is detected till the quadrotor has reached the goal position$\bm{p}_{G}$.}

\section{Deep Reinforcement Learning}

\subsection{Soft Actor-Critic Algorithm}
{Reward sparsity is a challenge for our task since the goal reward behind the gap is difficult to reach. For this, we {selected} Soft Actor-Critic (SAC) algorithm \cite{2}, which has a strong ability of exploration due to the entropy term $\mathcal{H}$ (refer to Eq.~(\ref{j_pi})). {Our preliminary experiments indicate that SAC converges faster than PPO \cite{PPO} and Deep Deterministic Policy Gradient (DDPG) \cite{DDPG}. Hence, SAC is chosen as the learning algorithm in this paper.}}
\begin{equation}
J(\pi)=\sum_{t=0}^{T} \mathbb{E}_{\left(\mathbf{s}_{t}, \mathbf{a}_{t}\right) \sim \rho_{\pi}}\left[r\left(\mathbf{s}_{t}, \mathbf{a}_{t}\right)+\alpha \mathcal{H}\left(\pi\left(\cdot |\mathbf{s}_{t}\right)\right)\right]
\label{j_pi}
\end{equation}

Where $r$ is the step reward, $\mathbf{s_{t}}$, $\mathbf{a_{t}}$ are the state and action in the time step $t$. $\alpha$ is a weight parameter that determines the importance of the entropy term {($\alpha$ is subsumed into the reward $r$ through scaling reward $r$ by $\alpha^{-1}$ \cite{2})}. The optimal policy $\pi_{\text {Maxent }}^{*}$ is given by (\ref{optim}):
\begin{equation}
\pi_{\mathrm{MaxEnt}}^{*}\left(\mathbf{a}_{t} | \mathbf{s}_{t}\right)=\exp \left(\frac{1}{\alpha}\left(Q_{\mathrm{soft}}^{*}\left(\mathbf{s}_{t}, \mathbf{a}_{t}\right)-V_{\mathrm{soft}}^{*}\left(\mathbf{s}_{t}\right)\right)\right)
\label{optim}
\end{equation}

The soft Q function $Q_{\text {soft}}^{*}$, soft V function $V_{\mathrm{soft}}^{*}$ are given by the soft learning framework, which are defined as Eq.~(\ref{sac_q}) and Eq.~(\ref{sac_v}). $\gamma$ is the reward discount factor.

\begin{equation}
\begin{aligned}
&Q_{\text {soft}}^{*}\left(\mathbf{s}_{t}, \mathbf{a}_{t}\right)=
r_{t}+\\ & \mathbb{E}_{\left(s_{t+1}, \ldots\right) \sim \rho_{\pi}}\left[\sum_{l=1}^{\infty} \gamma^{l}\left(r_{t+l}+\alpha \mathcal{H}\left(\pi_{\text {MaxEnt }}^{*}\left(\cdot |\mathbf{s}_{t+l}\right)\right)\right)\right] 
\end{aligned}
\label{sac_q}
\end{equation}

\begin{equation}
V_{\mathrm{soft}}^{*}\left(\mathbf{s}_{t}\right)=\alpha \log \int_{A}  \exp \left(\frac{1}{\alpha} Q_{\mathrm{soft}}^{*}\left(\mathbf{s}_{t}, \mathbf{a}^{\prime}\right)\right) d \mathbf{a}^{\prime}
\label{sac_v}
\end{equation}

\begin{figure}[thpb]
      \centering
	\includegraphics[scale=1.05]{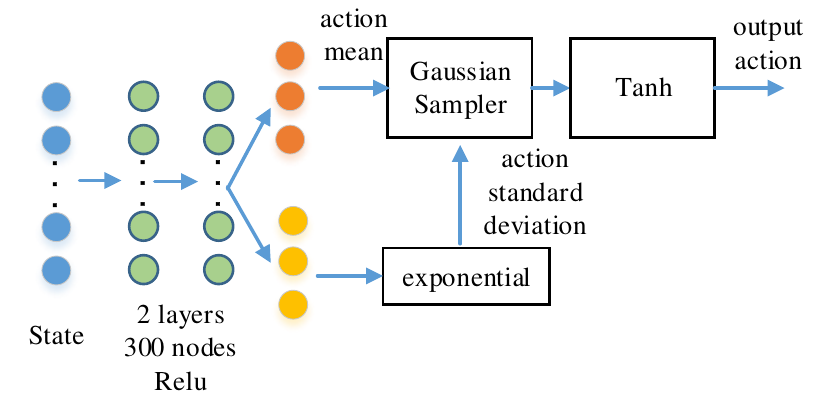}
        \caption{Architecture of the policy network that predicts the distribution of actions conditioned on the input state. Reparameterization trick is used for sampling actions.}
        \label{struct_policy_net}
\end{figure}

We approximate the policy with a neural network $\pi_{\phi}$. This policy network has 2 linear hidden layers with 256 neural units in each layer. Rectified Linear Unit (ReLU) activation function is used in all hidden layers. {We use reparameterization trick \cite{21} to sample actions i.e.  $\mathbf{a}_{t}=f_{\phi}\left(\epsilon_{t} ; \mathbf{s}_{t}\right)$, where $\epsilon_{t}$ is a noise signal sampled from a Gaussian distribution {defined by the network output}. We limit the action magnitude of each channel to (-1, 1) by a Tanh function. The overall network structure is given in Fig.~\ref{struct_policy_net}.}

$Q_{soft}^*$ function and $V_{soft} ^*$ are also approximated with neural networks $Q_{\theta}$ and $V_{\psi}$.  Both of the two networks contain 3 hidden layers with 300 neural units in each layer. To prevent the overestimation of Q value, {we follow the double-Q learning \cite{DQL1} \cite{DQL2} to approximate the $Q_{soft}^*$  with the minimum output of two parallel $Q$ networks.} 

We trained all these networks with Adam optimizer at a learning rate of $5 \times 10^{-4}$ and a batch size of 1024. {We identify that using a smaller learning rate (less than $1 \times 10^{-4}$) may lead to collapsed solution trajectories since it cannot follow the update speed of curriculum learning (refer to Sec.~\ref{Sec_CL}) while using a large learning rate (larger than $2 \times 10^{-3}$) may reduce training stability. The reward discount factor $\gamma$ is 0.99}. We initialize the weights of the output layer in $Q_{\theta}$  and $V_{\psi}$ as uniform values in $(-3\times10^{-3}, 3\times10^{-3})$, because we want to initialize the estimation of $Q_{soft}^*$ and $V_{soft}^*$ as roughly zero compared to the {relatively large episodic reward}. We believe this can {alleviate the bias in selecting initial actions and may accelerate the training}. 

\vspace{-3mm}
\subsection{Curriculum Learning} \label{Sec_CL}
{We incorporate our proposed curriculum learning framework to address the reward sparsity issue. Curriculum learning \cite{19} is a training technique that divides the training process into a sequence of subtasks with increased difficulty levels. {which is known to be able to improve the convergence by letting the agent learn on a simplified problem at the beginning stage \cite{FIREEVAQ}.}} 

We design a curriculum with two training phases. In phase 1, the gap's dimensions gradually reduce from 1.5m $\times$ 1m to 1m $\times$ 0.5m. This phase lasts for 100,000 episodes. We control the gap's dimension by increasing the difficulty factor $f_1$ with the episode $e_1$, as described in Eq.~(\ref{f1}). $w$ and $h$ are the width and height of the gap. 
\begin{equation}
\begin{array}{c}{f_{1}=\min \left(0.5 \sqrt{\frac{e_{1}}{10,000}}, 1.0\right)} \\ {w=1.5-0.5 \cdot f_{1}} \\ {h=1.0-0.5 \cdot f_{1}}\end{array}
\label{f1}
\end{equation}
{
In phase 2, we adjust the difficulty factor $f_2$ according to Eq.~(\ref{f2}). Phase 2 is used to refine the policy under the most difficult configuration. The phase 2 lasts for 500,000 steps in total. {\color{black} which shrinks the gap dimension from 1.0 $\times$ 0.5 to 0.6 $\times$ 0.3. }}
\begin{equation}
\begin{array}{c}{f_{2}=\min \left(0.5 \sqrt{\frac{e_{2}}{150,000}}, 1.0\right)} \\ {w=1.0-0.4 \cdot f_{2}} \\ {h=0.5-0.2 \cdot f_{2}}\end{array}
\label{f2}
\end{equation}
{
The best policy is chosen as the one with the maximized score $s=f_2 r^*$, where $r^*$ is the exponential moving average of the episode reward $r$ at episode $e_2$ ($r^*_{e_2+1} = 0.95r^*_{e_2} + 0.05r_{e_2}, r^*_0 = r_0$).}

{The curriculum learning changes the environmental configuration as the training proceeds. This means that the experience stored in the replay buffer may be obsolete.} Therefore, we limit the size of our replay buffer to 100,000 and discard old data when the replay buffer is full. {Empirically, the reward curve is stable when the replay buffer size varies from 10,000 to 500,000.}

\section{Sim2Real Transfer}
Discrepancies between the simulation and real quadrotors are non-negligible. {Therefore}, it is difficult to transfer the policy trained in {simulation} directly to real quadrotors. {To solve this problem, a wide variety of Sim2Real approaches have been proposed {\cite{10, 11, 12, 14}}. Nevertheless, most of these approaches need to utilize real-world data either in fine-tuning stage or in training stage. However, acquiring real-world data is challenging in our case (discussed in Sec.~\ref{sec:discussion}). To solve this problem, we developed a control framework that can enhance generalization without utilizing real-world data. }

\vspace{-3mm}
\subsection{Simulation to Real Transfer Framework}

\begin{figure}[htb]
    \centering
	\includegraphics[width=\linewidth]{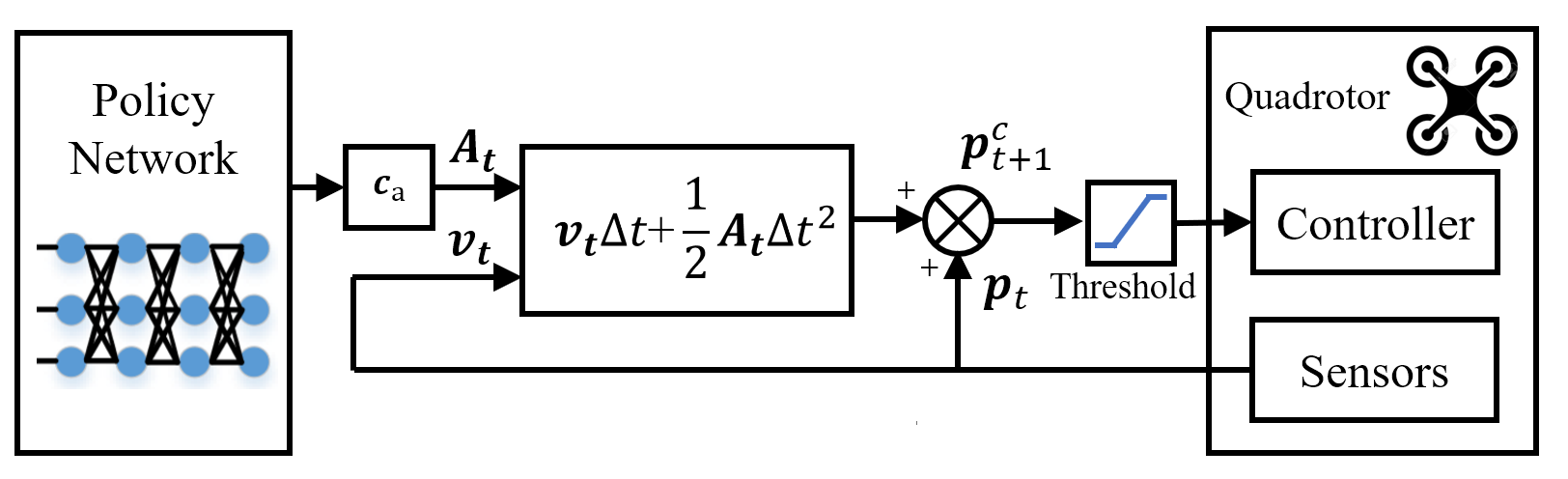}
    \caption{{\color{black}Our proposed Sim2Real transfer framework. Position command $\bm{p}^{c}_{t+1}$ at time step $t+1$ is calculated using the acceleration command $\bm{A}_t$ and positional and velocity feedback $\bm{p}_{t}$, $\bm{v}_{t}$ at time step $t$}}
    \label{transfer_overview}
    \vspace{-5mm}
\end{figure}

An overview of our framework is shown in Fig.~\ref{transfer_overview}. The proposed framework is incorporated both in training and testing. Here we define the linear and angular acceleration command as $\bm{A_{t}}$. $\bm{A_{t}}$ is then converted into an incremental positional displacement starting from the current position $\bm{p}_t$. 

{Let $\bm{p}_{t}$, $\bm{v}_{t}$ denote the position and velocity of the quadrotor at time step $t$, respectively. We propose to design the position command as follows:
\begin{equation}
\bm{p}^{c}_{t+1} = \bm{p}_{t}+ \bm{v}_{t} \Delta t+ \frac{1}{2}\bm{A}_{t} \Delta t^2
\label{motion_xddot}
\end{equation}
where $\bm{p}^{c}_{t+1}$ denotes the position command for the next time step ($t+1$), $\Delta t$ denotes the time interval between the two time steps.

$\bm{p}^{c}_{t+1}$ will be sent to the position controller for execution. The velocity $\bm{v}_{t}$ and position $\bm{p}_{t}$ are measured by sensors in real time. In our implementation, the policy network's output $\mathbf{a}$ is limited to $(-1, 1)$ by a $tanh$ function. To convert this output value back to $\bm{A}_t$, a scaling parameter $\bm{c}_a$ is used following the output of the policy network. For $\bm{c}_a$, we set $40 \ rad/s^2$ for angular channels and $12\ m/s^2$ for the altitude channel.
}

Our approach theoretically can work with continuous output, model-free reinforcement learning algorithms other than SAC, since it doesn't require
any modification of the existing reinforcement learning architectures.

\vspace{-3mm}
\subsection{Randomization}

Randomization is an effective way to enhance the success rate of Sim2Real transfer \cite{14, 8}. In our training, we use two types of randomization: (1) Observation noise that represents the uncertainty of sensors. (2) Dynamics randomization that represents the model inaccuracy.

Noise (1) is modeled as additive noise sampled from Gaussian distributions $\mathcal{N}(\mu,\,\sigma^{2})$. The mean value $\mu$ of noise is zero. The standard {deviation $\sigma$ is given in Table.~\ref{env_rand}. The initial state of the quadrotor is randomized by generating from zero-mean Gaussian distributions with standard deviations given in Table.~\ref{init_rand}, which enables to plan trajectories starting from a wide region rather than only from the origin.}

\renewcommand{\arraystretch}{1.15}
\begin{table}[h]
\caption{Environmental randomization}
\label{env_rand}
\centering
\begin{tabular}{lllll}
\hline
&  position & angle & linear velocity  & angular velocity \\
 & $p_x$, $p_y$, $p_z$ & $\phi$, $\theta$, $\psi$ & $v_x$, $v_y$, $v_z$ & $\omega_{x}$, $\omega_{y}$, $\omega_{z}$\\
\hline
$\sigma$ & 0.002 m & 0.01 rad & 0.05 m/s & 0.05 rad/s\\
\hline
\end{tabular}
\vspace{-2mm}
\end{table}
\begin{table}[h]
\caption{Initialization randomization}
\label{init_rand}
\centering
\begin{tabular}{lllll}
\hline
&  Initial linear velocity  &Initial angular velocity  & Initial position   \\
 & $v_{x0}$, $v_{y0}$ & $\omega_{x0}$, $\omega_{y0}$, $\omega_{z0}$ & $p_{x0}$, $p_{y0}$, $p_{z0}$  \\ \hline
$\sigma$ &  $0.01 m/s  $ & $ 0.01 rad/s   $ & \begin{tabular}[c]{@{}l@{}}$p_{x0}, p_{y0} : 0.5 m  $\\ $p_{z0} : 0.2 m $\end{tabular} \\
\hline
\end{tabular}
\vspace{-2mm}
\end{table}

{The dynamics randomization} aims at pushing the learning algorithm to generalize on a wide range of quadrotor parameters. {For this, we leverage additive zero mean Gaussian distributions, with} standard derivation $\sigma$ given in Table.~\ref{dyn_rand}.
\begin{table}[h]
\caption{Dynamics randomization}
\label{dyn_rand}
\centering
\begin{tabular}{lllll}
\hline
&  rotational inertia  & motor's max thrust   \\
 & $I=[I_{xx}, I_{yy}, I_{zz}]^T$ & $T_{max}=[T_{max1},...,T_{max4}]^T$ \\ \hline
$\sigma$ &  $0.15 I $ & $0.05T_{max} $ \\
\hline
\end{tabular}
\vspace{-5mm}
\end{table}

\subsection{Traversing through gaps with various dimensions} \label{EXP:Traversing Success Rate}
{To demonstrate the feasibility of our approach, we firstly} evaluate the traversing success rate of our policy with various gap dimensions. The dimension of our quadrotor is 0.47m $\times$ 0.47m $\times$ 0.23m. The dynamics parameters of the quadrotor are $m=1.2 \ kg$, {total thrust $f_t=19.6 \ N$}, rotational inertia $I_{xx}=I_{yy}=0.007\ kg\cdot m^{2}$, $I_{zz}=0.014 kg\cdot m^{2}$, thrust coefficient $C_{T}=6 \times 10^{-6} \ N/(rad/s)^{2}$ and torque coefficient $C_{M}=8 \times 10^{-8} \ N\cdot m/(rad/s)^{2}$, which is consistent with our real quadrotor. Both the training and testing stages are conducted in the simulation we built, which runs on a laptop with intel i7-8750H CPU and Nvidia GTX 1060 GPU. The tilted angle is fixed to 20 degrees in both training and testing as an example. We evaluate our approach on a wide variety of gap dimensions, with 1,000 episodes evaluated per experiment. The success rate is shown in Table \ref{sucess_rate}.
\begin{table}[thpb]

\caption{{Evaluation of the policy in simulation. We demonstrate the success rate (in \%) for various gap dimensions (width \& height, in meters)} }
\vspace{-1mm}
\label{sucess_rate}
\begin{center}
\begin{tabular}{llllll}
\hline
\multicolumn{1}{|l|}{\diagbox{width}{height}} & \multicolumn{1}{l|}{0.38} & \multicolumn{1}{l|}{0.36} & \multicolumn{1}{l|}{0.34} & \multicolumn{1}{l|}{0.32} & \multicolumn{1}{l|}{0.30} \\ \hline
\multicolumn{1}{|l|}{1.0} & \multicolumn{1}{l|}{95.1\%} & \multicolumn{1}{l|}{93.0\%} & \multicolumn{1}{l|}{86.4\%} & \multicolumn{1}{l|}{70.5\%} & \multicolumn{1}{l|}{49.2\%} \\ \hline
\multicolumn{1}{|l|}{0.9} & \multicolumn{1}{l|}{90.0\%} & \multicolumn{1}{l|}{88.5\%} & \multicolumn{1}{l|}{83.5\%} & \multicolumn{1}{l|}{70.8\%} & \multicolumn{1}{l|}{46.6\%} \\ \hline
\multicolumn{1}{|l|}{0.8} & \multicolumn{1}{l|}{78.4\%} & \multicolumn{1}{l|}{75.8\%} & \multicolumn{1}{l|}{72.0\%} & \multicolumn{1}{l|}{58.6\%} & \multicolumn{1}{l|}{40.9\%} \\ \hline
\multicolumn{1}{|l|}{0.7} & \multicolumn{1}{l|}{45.6\%} & \multicolumn{1}{l|}{44.6\%} & \multicolumn{1}{l|}{42.8\%} & \multicolumn{1}{l|}{36.3\%} & \multicolumn{1}{l|}{24.0\%} \\ \hline
\multicolumn{1}{|l|}{0.6} & \multicolumn{1}{l|}{14.7\%} & \multicolumn{1}{l|}{12.6\%} & \multicolumn{1}{l|}{13.8\%} & \multicolumn{1}{l|}{11.6\%} & \multicolumn{1}{l|}{7.9\%} \\ \hline
\end{tabular}
\end{center}
\vspace{-3mm}
\end{table}

\begin{figure*}\centering%
\captionsetup{justification=centering}
\centering
\begin{minipage}[b]{0.23\linewidth}
\subfigure[Roll command and response.]{\includegraphics[width=4.3cm, angle=0]{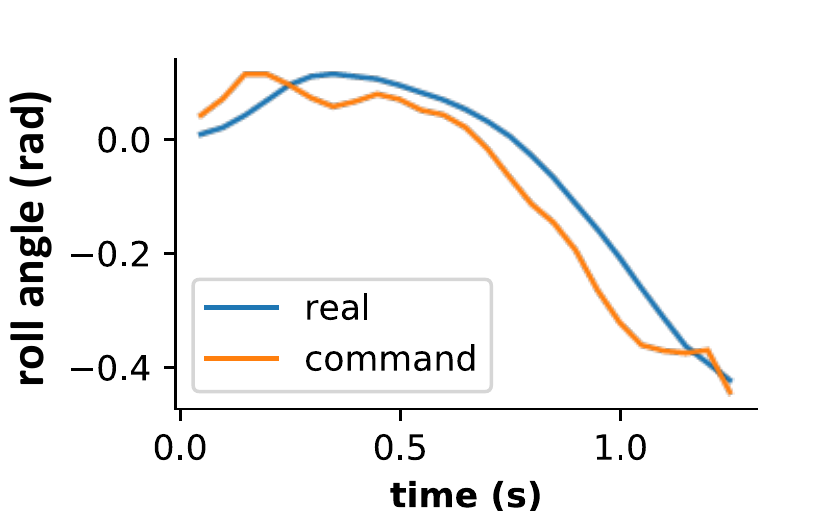}}
\end{minipage}
\begin{minipage}[b]{0.23\linewidth}
\subfigure[Pitch command and response]{\includegraphics[width=4.3cm, angle=0]{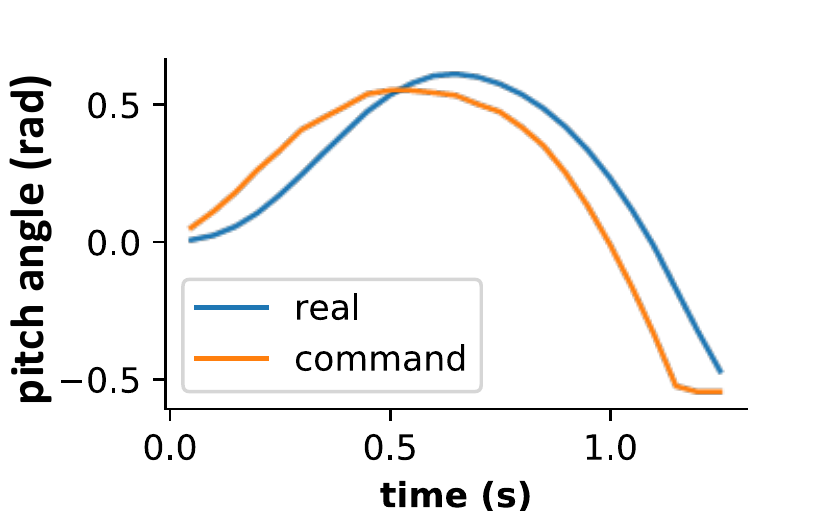}}
\end{minipage}
\begin{minipage}[b]{0.23\linewidth}
\subfigure[Altitude command and response]{\includegraphics[width=4.3cm, angle=0]{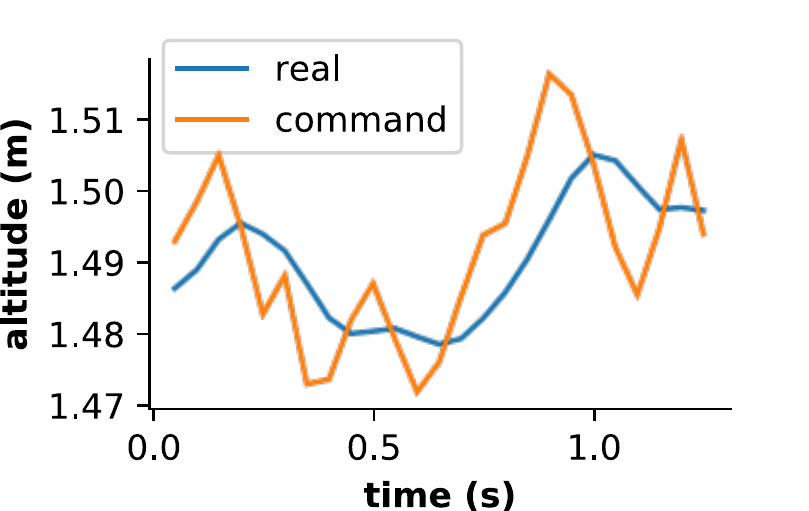}}
\end{minipage}
\begin{minipage}[b]{0.23\linewidth}
\subfigure[Trajectory of traversing]{\includegraphics[scale=0.2, angle=0]{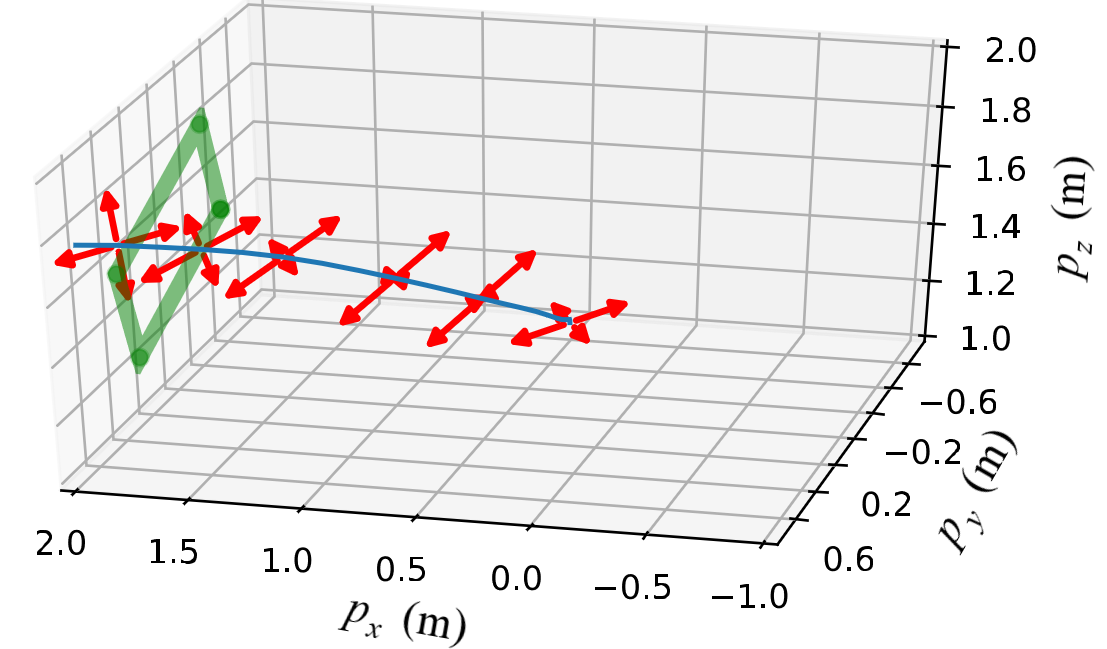}}
\end{minipage}
\vspace{-1mm}
\caption{Experimental data for traversing through a 20 degree tilted gap. Quadrotor attitude and altitude data are shown in (a), (b), and (c). (d) shows a trajectory that passes through the narrow gap successfully in our simulation.}
\label{sim_exp}
\vspace{-3mm}
\end{figure*}

We demonstrate the learned policy by showing plots of the altitude and attitude data (Fig.~\ref{sim_exp}). The pitch angle gradually increases to obtain a fast dashing speed. Then it gradually decreases because a large pitch angle may increase the chance of collision. The quadrotor finally takes advantage of the inertial velocity for the hole-traversing.

\vspace{-3mm}
\subsection{Real World Experimental Configuration}
\label{s:63}
{To show the feasibility of our proposed Sim2Real method,} we then test our approach on a real F330 quadrotor. The parameters from model identification are the same as the counterparts in Sec.~\ref{EXP:Traversing Success Rate}. The width of the gap is 0.7m and the height is 0.36m. The tilt angle is 20 degrees. We set the quadrotor’s absolute maximum roll/pitch angle as 0.55 rad (about 31.5 degrees) to prevent losing altitude due to limited motor thrust.

\begin{figure}[thpb]
        \centering
        \includegraphics[width=0.9\linewidth]{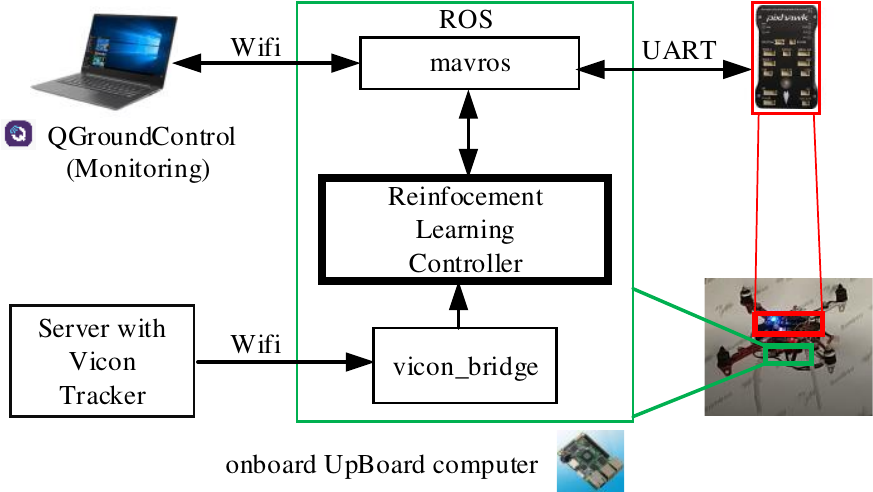}
        \caption{The experimental configuration of our real-world experiment. The reinforcement learning controller is on an onboard Upboard computer. A Pixhawk module is used for flight control. Vicon tracker is used for position feedback.}
        \label{exp_conf}
\end{figure}

We utilize Vicon mocap system to provide the position and velocity feedback. The whole reinforcement learning framework was running on an onboard Upboard computer with the Robot Operating System (ROS). The system structure is shown in Fig.~\ref{exp_conf}. The positional channels (outer loops) are controlled at 50 Hz while the attitude is controlled by the onboard Pixhawk controller at 250 Hz rate. Our code is released at: \url{https://github.com/arclab-hku/reinforcement_learning}.

\begin{figure*}\centering%
\subfigure[]{\includegraphics[width=4.2cm, angle=0]{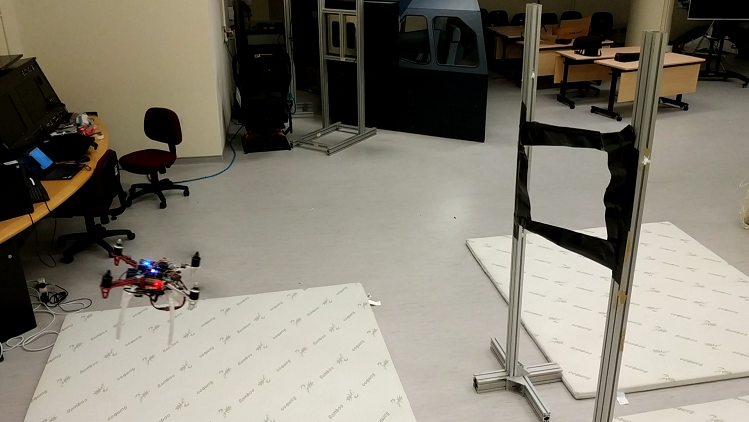}}
\subfigure[]{\includegraphics[width=4.2cm, angle=0]{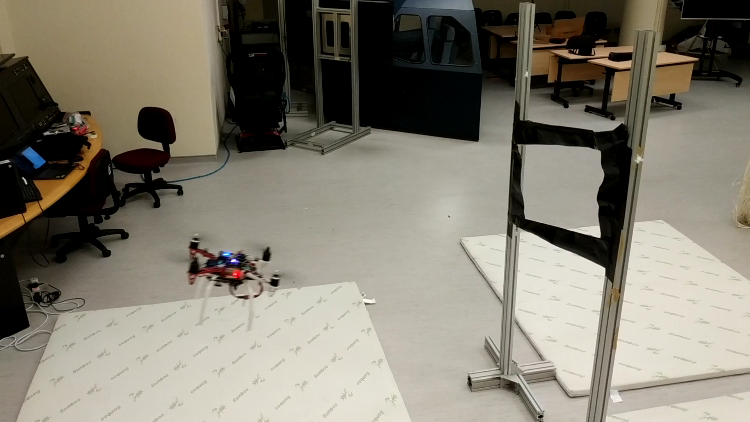}}
\subfigure[]{\includegraphics[width=4.2cm, angle=0]{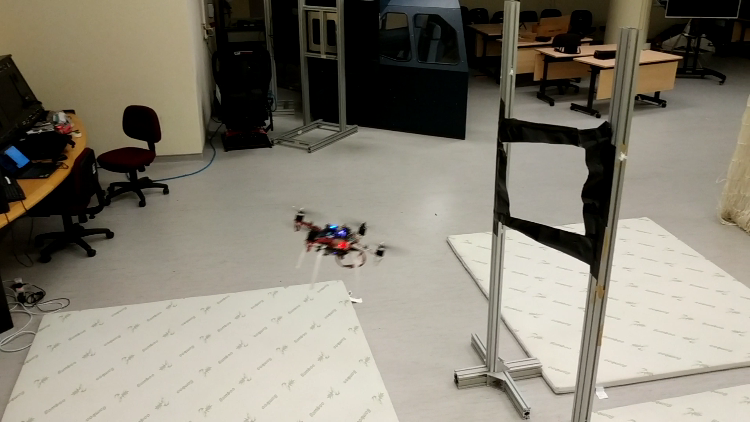}}
\subfigure[]{\includegraphics[width=4.2cm, angle=0]{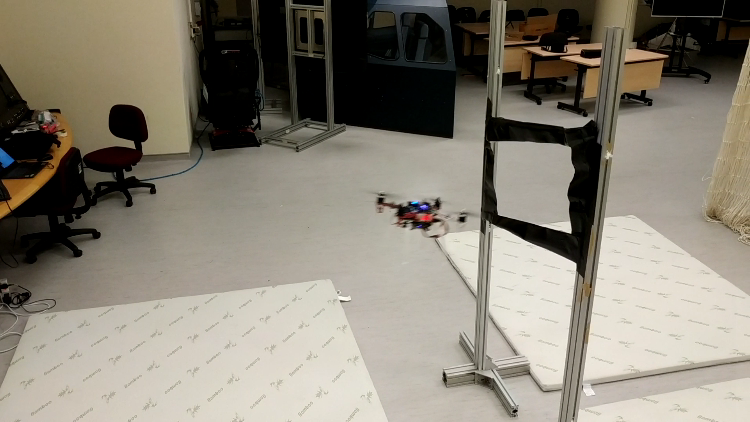}}
\subfigure[]{\includegraphics[width=4.2cm, angle=0]{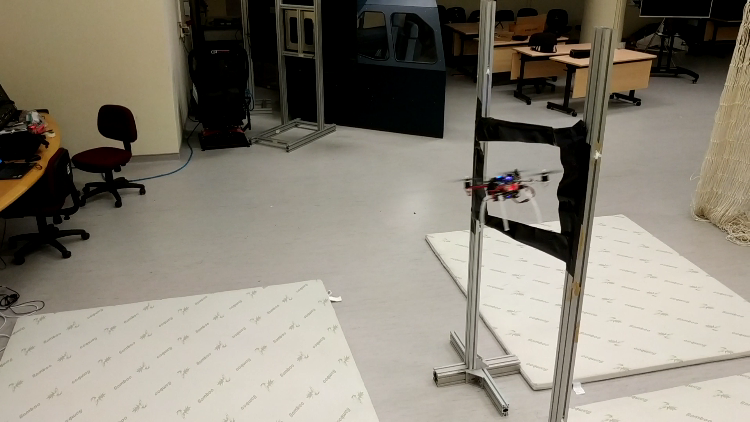}}
\subfigure[]{\includegraphics[width=4.2cm, angle=0]{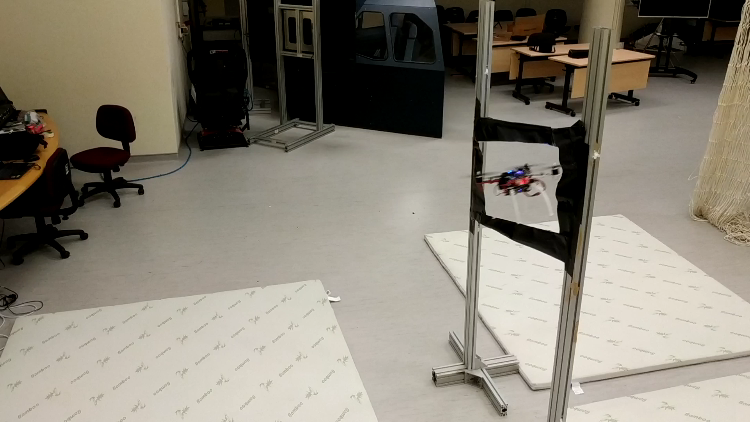}}
\subfigure[]{\includegraphics[width=4.2cm, angle=0]{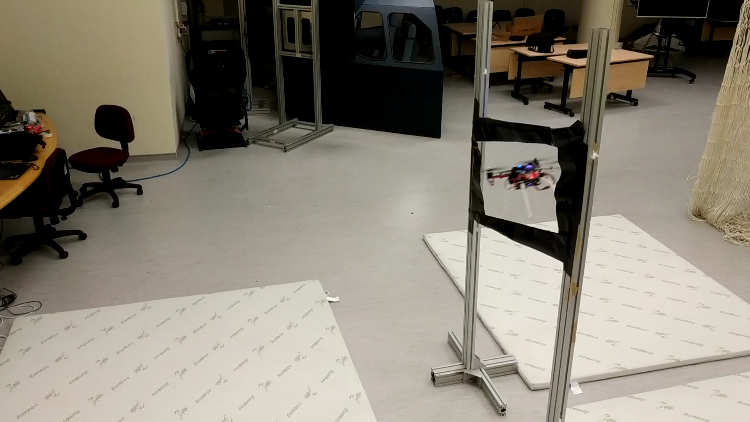}}
\subfigure[]{\includegraphics[width=4.2cm, angle=0]{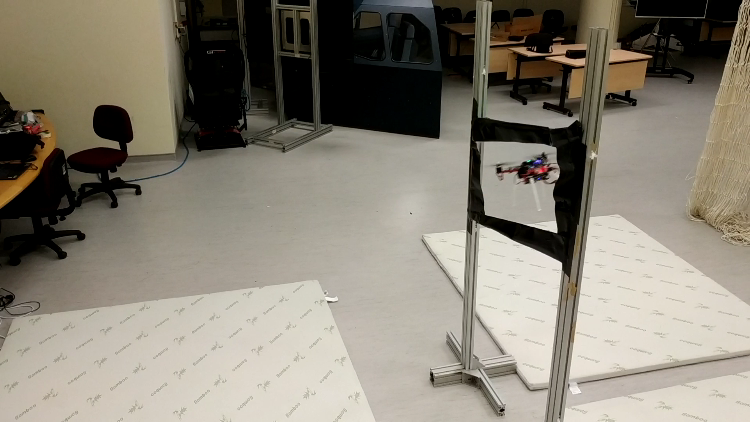}}
\vspace{-2mm}
\caption{Snapshots of the motions performed during the gap-traversing task.}
\label{video_seg}  
\vspace{-6mm}
\end{figure*}

We demonstrate the results of real world experiment. We conducted 37 trials of experiments. 15 of them are successful, which takes up about 40.54\%. This success rate is close to 44.6\% we achieved in the simulation. The traversing snapshots are shown in Fig.~\ref{video_seg}. The video can be found at \url{https://youtu.be/gfAfFnjN18A}.

\begin{figure*}\centering%
\captionsetup{justification=centering}
\centering
\begin{minipage}[b]{0.23\linewidth}
\subfigure[Roll command and response.]{\includegraphics[width=4.3cm, angle=0]{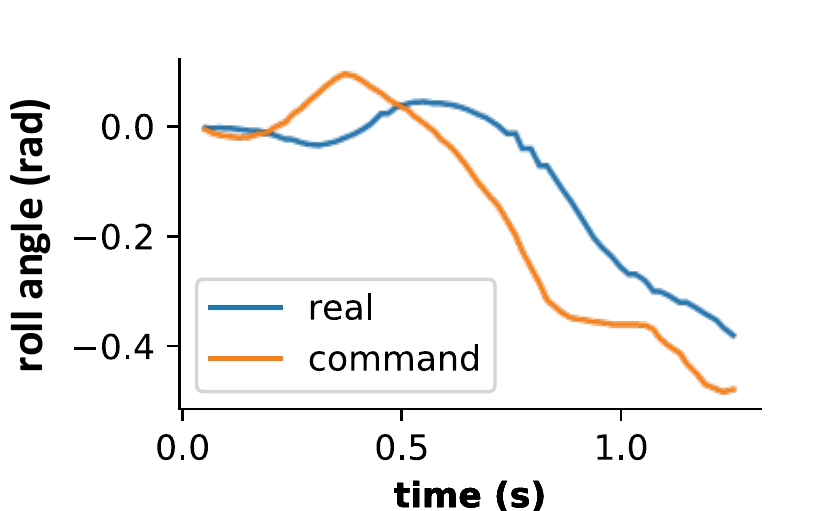}}
\end{minipage}
\begin{minipage}[b]{0.23\linewidth}
\subfigure[Pitch command and response.]{\includegraphics[width=4.3cm, angle=0]{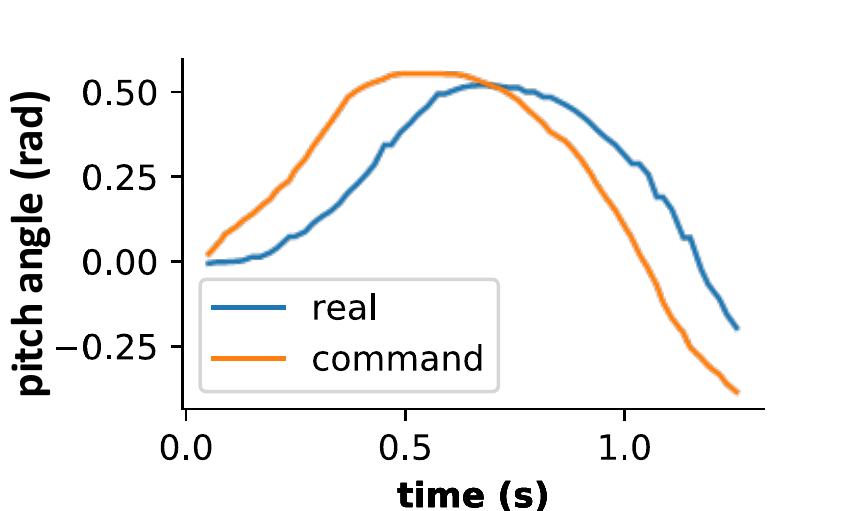}}
\end{minipage}
\begin{minipage}[b]{0.23\linewidth}
\subfigure[Altitude command and response.]{\includegraphics[width=4.3cm, angle=0]{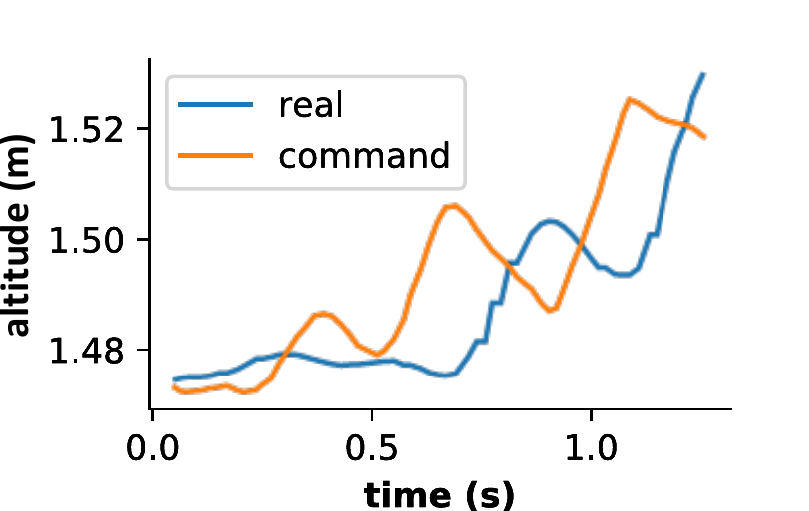}}
\end{minipage}
\begin{minipage}[b]{0.23\linewidth}
\subfigure[Trajectory of traversing]{\includegraphics[scale=0.2, angle=0]{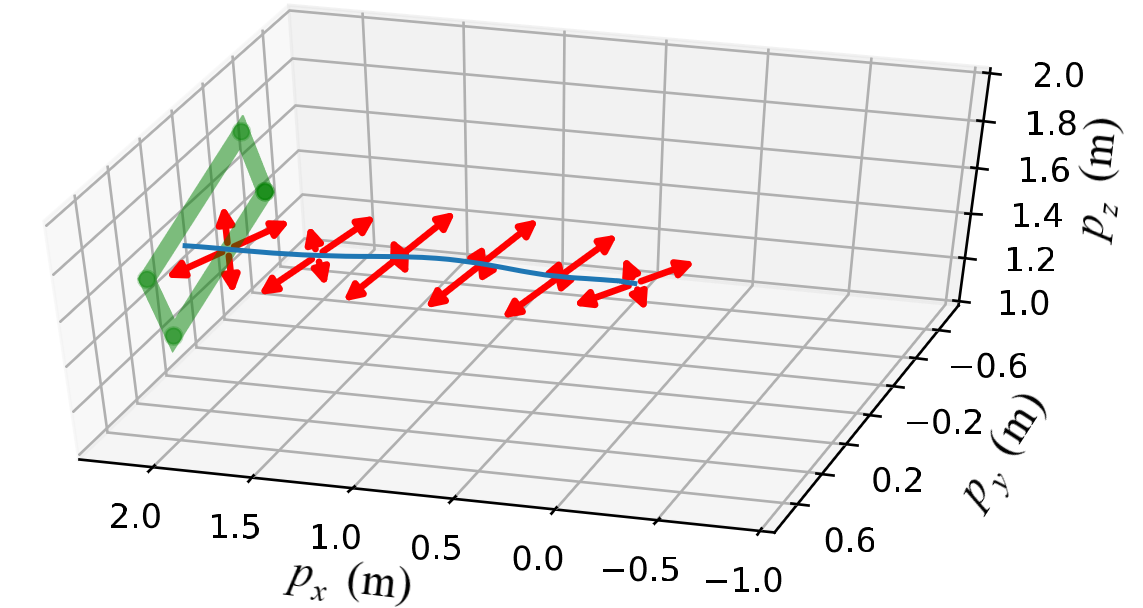}}
\end{minipage}
\vspace{-3mm}
\caption{Quadrotor states in a real-world experiment. (a), (b) and (c) show the command and response. (d) is the recorded trajectory that passed the narrow gap successfully.}
\label{sim_real}   
\vspace{-4mm}
\end{figure*}

The key state variables in actual flights are demonstrated in Fig.~\ref{sim_real}. It can be observed that the action pattern closely matches the simulated counterparts. This demonstrates that our Sim2Real framework can effectively transfer the policy from simulation to a real quadrotor.

\begin{figure}[htb]
 \centering
 \includegraphics[scale=0.5]{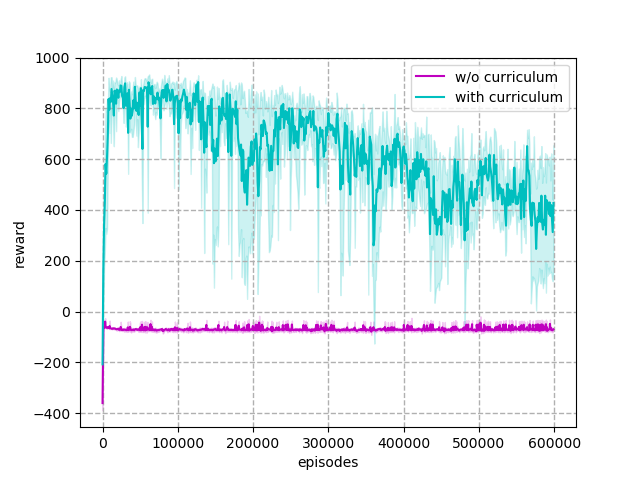}
  \caption{Comparison of smoothed episodic reward of two training configurations: 1) With curriculum learning 2) Without curriculum learning. {Both configurations have the same gap dimension (0.6x0.3) at the 600,000th episode. But only the former case can find the solution trajectory reliably.}}
  \label{curriculum}
  \vspace{-7mm}
\end{figure}

{
\vspace{-2mm}
\subsection{Performance without curriculum learning}
We demonstrate the smoothed episodic reward $r'$ (smoothed by $r'_{t+1} = 0.995r'_{t}+0.005r_{t+1}$ ) in Fig.~\ref{curriculum}, {with 95\% confidence intervals}. The cyan curve corresponds to the results with curriculum learning enabled, while the pink curve corresponds to the result with curriculum learning removed. Benefits from the curriculum learning, the cyan curve can maintain a high reward level during the whole training process. In comparison, the pink curve shows that the agent is not able to find the goal reward when the curriculum learning is removed, proving that curriculum learning can both improve the learning speed and stability.
}

\vspace{-2mm}
\subsection{Performance without Sim2Real transfer framework}
We find it intractable to transfer a policy that directly exerts control on the attitude and altitude channels without using our proposed Sim2Real framework. For safety considerations, we only tested this transfer in simulation: we trained the policy using the simulated dynamics model and then transferred it to a quadrotor model controlled by PX4 firmware in Gazebo. No successful trajectory is achieved with a total number of 30 rollouts while at the same scenario we can achieve a success rate of 44.6\% in the simulation using the proposed framework. 

\begin{figure*}\centering%
\captionsetup{justification=centering}
\centering
\begin{minipage}[b]{0.23\linewidth}
\subfigure[Roll command and response.]{\includegraphics[width=4.3cm, angle=0]{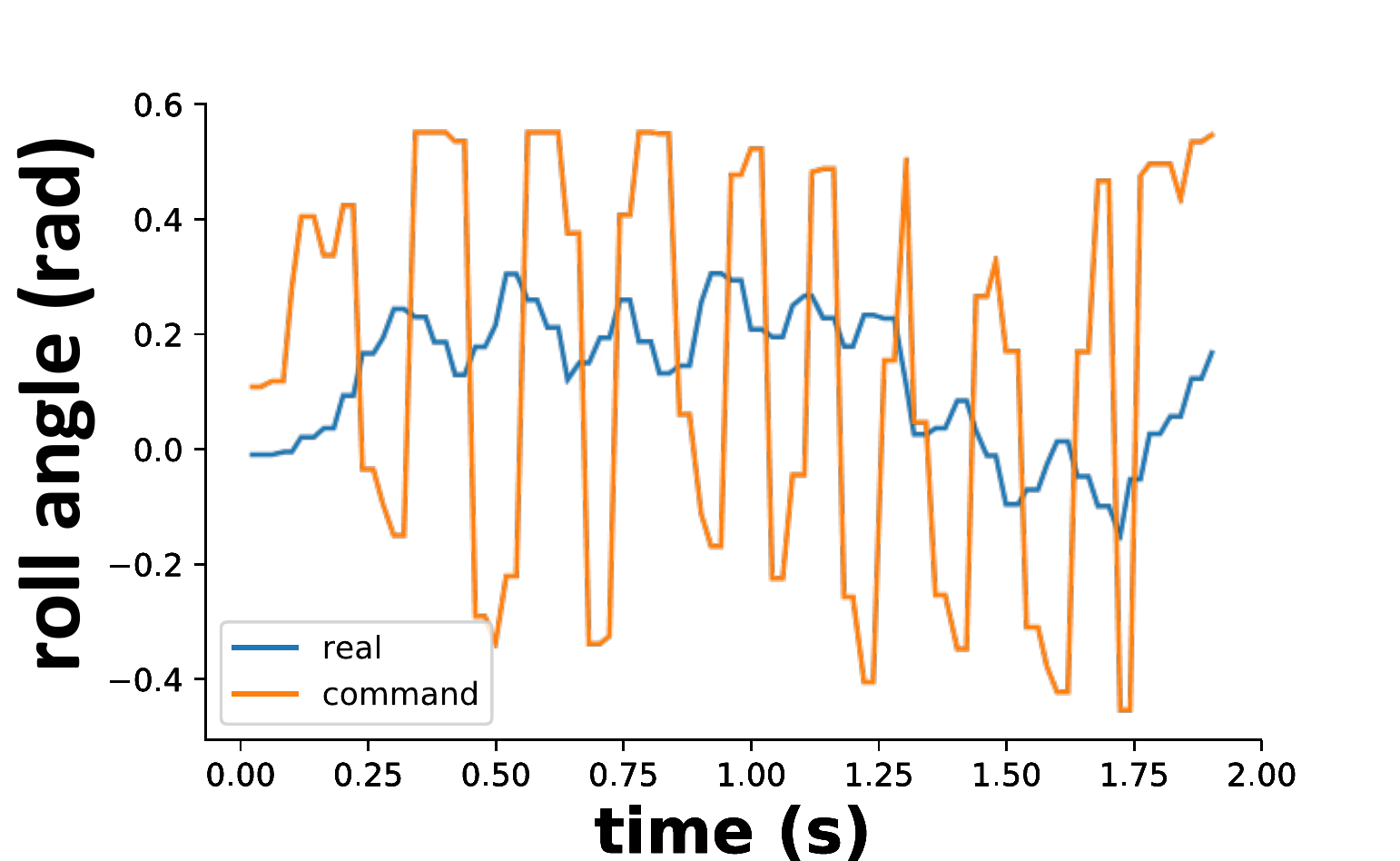}}
\end{minipage}
\begin{minipage}[b]{0.23\linewidth}
\subfigure[Pitch command and response.]{\includegraphics[width=4.3cm, angle=0]{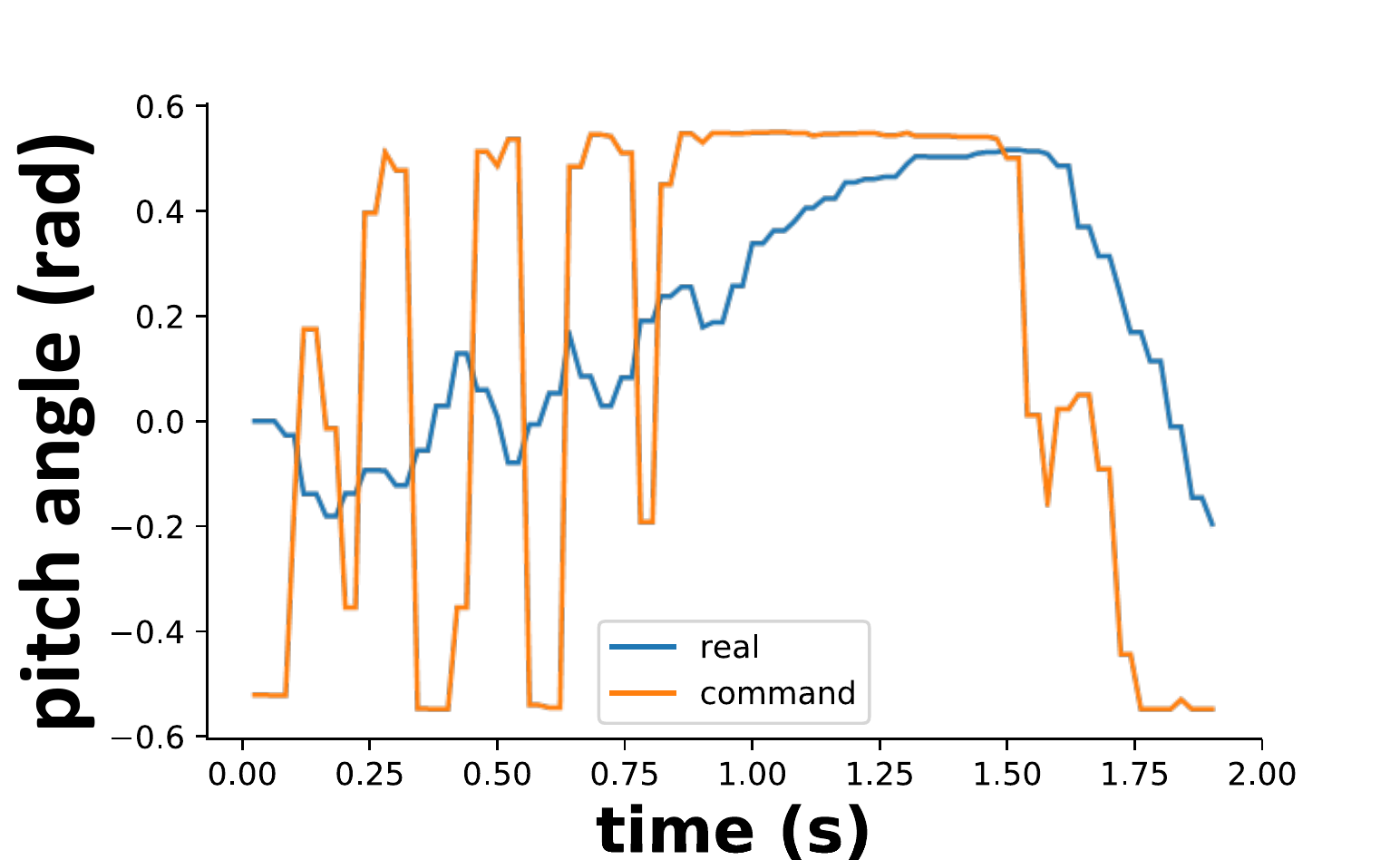}}
\end{minipage}
\begin{minipage}[b]{0.23\linewidth}
\subfigure[Altitude command and response.]{\includegraphics[width=4.3cm, angle=0]{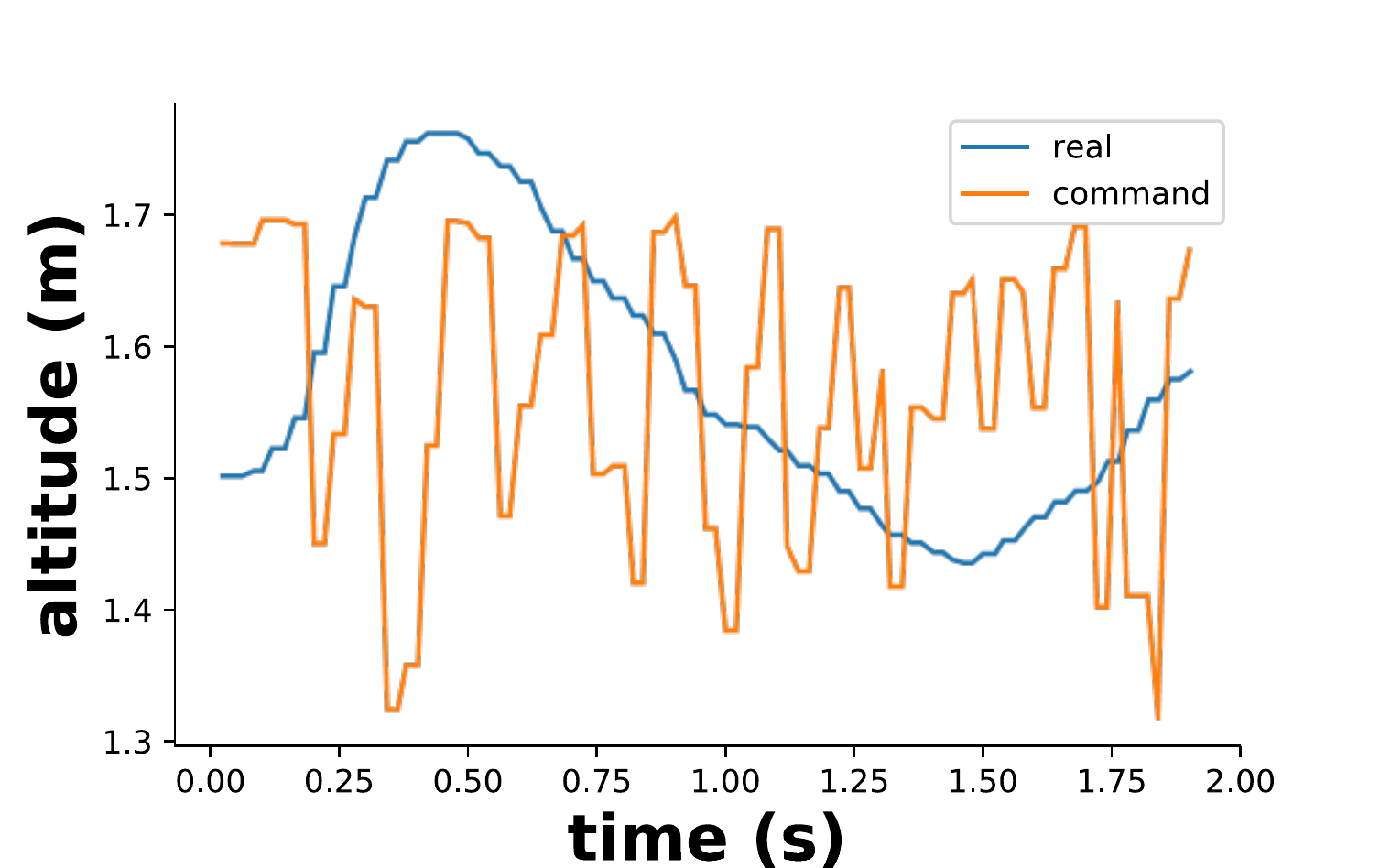}}
\end{minipage}
\begin{minipage}[b]{0.23\linewidth}
\subfigure[Trajectory of traversing]{\includegraphics[scale=0.2, angle=0]{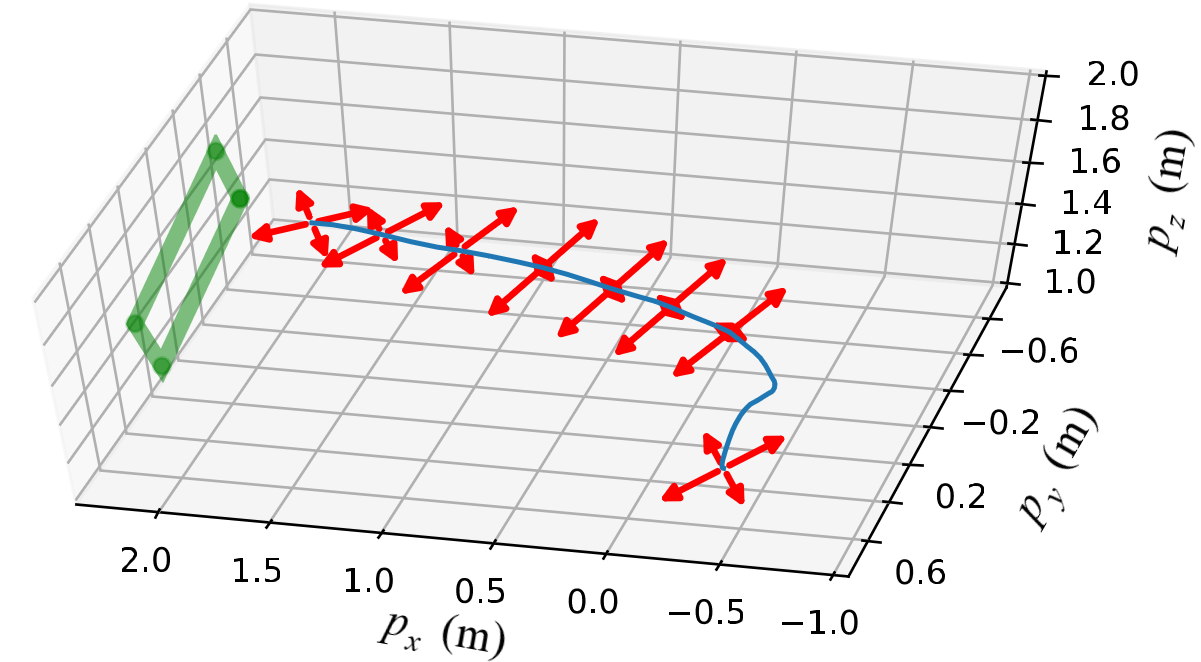}}
\end{minipage}
\vspace{-2mm}
\caption{Quadrotor response from a failure trajectory without using our proposed sim2real transfer framework in Gazebo environment. (a), (b) and (c) show the command and response. The commands are oscillatory, which leads to task failure. (d) is the corresponding recorded trajectory. The quadrotor collided with the wall.}
\label{trans_fail}  
\vspace{-3mm}
\end{figure*}

A planning result in Gazebo is shown in Fig.~ \ref{trans_fail}. It is seen that the attitude and altitude response is oscillatory, making it difficult to track the commands.
\vspace{-2mm}
\section{Discussions} \label{sec:discussion}
\subsection{Other Sim2Real Approaches}
Other recent proposed approaches mainly include: (1) learn a model of inverse dynamics that can predict required actions {directly in the target domain} \cite{10}. (2) learn an adaptive policy that can be fine-tuned by real-world data \cite{12} \cite{15}. Unfortunately, none of these approaches is effective in our system.

(1) We have tried an inverse dynamic model as an attempt of the Sim2Real transfer (refer to \cite{10}). However, it is intractable to fit an accurate global model or local models around aggressive trajectories, because a real quadrotor is fragile and therefore intensive data sampling around aggressive trajectories is not feasible. We have tried to use Ornstein-Uhlenbeck noise for model identification, but the noise magnitude should also be limited due to safety considerations. Hence, it is hard to bridge the data distribution gap between the identification phase and the validation phase.

(2) We seek an antidote in fast adaptive meta-learning by applying the Reptile algorithm \cite{15}. By generating 1,000 quadrotors with dynamics randomization in our simulation, we intended to find a well-initialized model, and then fine-tune the model by the data acquired from the target domain. We use a Gazebo environment for the experiment. Using 5 shots of training, we achieve at most 3 successful rollouts out of 30 rollouts in total, which is a mundane performance compared to 10 successful rollouts achieved by our Sim2Real transfer framework. 

(3) {Other approaches that require real world data for domain transfer such as \cite{ADR} are also intractable to be applied due to the difficulty of sampling a large number of aggressive trajectories from the real-world. This is because almost any failure trials would damage the quadrotor e.g. break propellers.}

\subsection{Failure pattern analysis}
We aim to get the best performance on a real-world quadrotor rather than on the simulated counterparts. We can achieve more than 90\% success rate in our simulation if we decrease the noise injected for Sim2Real, but it will degenerate the performance on a real quadrotor.

Failures are caused by 1) inappropriate timing to start tilting, which implies that inaccurate decisions can still be made by the reinforcement learning agent.  2) inaccurate tracking of the altitude. The error in the altitude channel cannot be reduced swiftly once emerges, because the time constant in the altitude control channel is larger than the counterparts in attitude channels. Note that the controller only has fractions of a second for stabilization because the peak dashing speed of our quadrotor can be more than $3m/s$. {A better altitude control algorithm that has a faster control response (such as the incremental nonlinear dynamic inversion in \cite{13b}) may contribute to a higher success rate. }

{
\subsection{Generalizability}
The proposed Sim2Real transfer framework, which does not need accurate parameters of the real quadrotor, makes the proposed approach less dependent on the model of the quadrotor and easy to be generalized. Because of this, the trained network in simulation can be successfully applied to the real quadrotor without training on the real data and achieves a similar success rate as in simulation. This demonstrates the generalizability of the proposed approach. The proposed Sim2Real transfer framework can be generalized to systems with similar dynamics as the quadrotor.

}

\subsection{Limitation}

{For performing aggressive flights using reinforcement learning approaches, angle and rate limits can be violated. One approach to attenuate this issue is to design reward functions which penalize the actions that violate the rate limit. This approach can attenuate the issue but cannot eradicate it. Our proposed Sim2Real framework takes a further step by always keeping the rate limit within its maximum range. However, the maximum rate limit is a function of quadrotor state. Simply using a constant rate limit value would be harmful when generalizing to larger tilt angles. 
}

\section{Conclusion}

{
We proposed a novel deep learning framework which enables the quadrotor to pass through narrow gaps without training using real-world data. Two key challenges were addressed: 1) the sparse reward issue was solved by designing a curriculum learning framework, and 2) the Sim2Real transfer issue was addressed by proposing a novel framework which does not depend on the model parameters. Experimental results showed that the trained policy can achieve a similar success rate when applied to the real quadrotor without additional training. Future work would be to extend our work to scenarios with larger tilted angles using a more dexterous quadrotor, and {to feed the gap's tilt angle to the network input, which can facilitate our proposed method to address varying tilting angles without the need to re-train the model. }
} 
\ifCLASSOPTIONcaptionsoff
  \newpage
\fi

\bibliographystyle{IEEEtran}
\bibliography{mybib}

\begin{thebibliography}{10}
\providecommand{\url}[1]{#1}
\csname url@samestyle\endcsname
\providecommand{\newblock}{\relax}
\providecommand{\bibinfo}[2]{#2}
\providecommand{\BIBentrySTDinterwordspacing}{\spaceskip=0pt\relax}
\providecommand{\BIBentryALTinterwordstretchfactor}{4}
\providecommand{\BIBentryALTinterwordspacing}{\spaceskip=\fontdimen2\font plus
\BIBentryALTinterwordstretchfactor\fontdimen3\font minus
  \fontdimen4\font\relax}
\providecommand{\BIBforeignlanguage}[2]{{%
\expandafter\ifx\csname l@#1\endcsname\relax
\typeout{** WARNING: IEEEtran.bst: No hyphenation pattern has been}%
\typeout{** loaded for the language `#1'. Using the pattern for}%
\typeout{** the default language instead.}%
\else
\language=\csname l@#1\endcsname
\fi
#2}}
\providecommand{\BIBdecl}{\relax}
\BIBdecl

\bibitem{3}
D.~Falanga, E.~Mueggler, M.~Faessler, and D.~Scaramuzza, ``Aggressive quadrotor
  flight through narrow gaps with onboard sensing and computing using active
  vision,'' in \emph{2017 IEEE international conference on robotics and
  automation (ICRA)}.\hskip 1em plus 0.5em minus 0.4em\relax IEEE, 2017, pp.
  5774--5781.

\bibitem{4}
G.~Loianno, C.~Brunner, G.~McGrath, and V.~Kumar, ``Estimation, control, and
  planning for aggressive flight with a small quadrotor with a single camera
  and imu,'' \emph{IEEE Robotics and Automation Letters}, vol.~2, no.~2, pp.
  404--411, 2016.

\bibitem{5}
T.~Zhang, G.~Kahn, S.~Levine, and P.~Abbeel, ``Learning deep control policies
  for autonomous aerial vehicles with mpc-guided policy search,'' in \emph{2016
  IEEE international conference on robotics and automation (ICRA)}.\hskip 1em
  plus 0.5em minus 0.4em\relax IEEE, 2016, pp. 528--535.

\bibitem{6}
J.~Hwangbo, I.~Sa, R.~Siegwart, and M.~Hutter, ``Control of a quadrotor with
  reinforcement learning,'' \emph{IEEE Robotics and Automation Letters},
  vol.~2, no.~4, pp. 2096--2103, 2017.

\bibitem{16}
A.~Molchanov, T.~Chen, W.~H{\"o}nig, J.~A. Preiss, N.~Ayanian, and G.~S.
  Sukhatme, ``Sim-to-(multi)-real: Transfer of low-level robust control
  policies to multiple quadrotors,'' \emph{arXiv preprint arXiv:1903.04628},
  2019.

\bibitem{17}
N.~O. Lambert, D.~S. Drew, J.~Yaconelli, S.~Levine, R.~Calandra, and K.~S.
  Pister, ``Low-level control of a quadrotor with deep model-based
  reinforcement learning,'' \emph{IEEE Robotics and Automation Letters},
  vol.~4, no.~4, pp. 4224--4230, 2019.

\bibitem{9}
S.~Li, T.~Liu, C.~Zhang, D.-Y. Yeung, and S.~Shen, ``Learning unmanned aerial
  vehicle control for autonomous target following,'' \emph{arXiv preprint
  arXiv:1709.08233}, 2017.

\bibitem{Mannucci2018}
T.~Mannucci, E.-J. van Kampen, C.~de~Visser, and Q.~Chu, ``Safe exploration
  algorithms for reinforcement learning controllers,'' \emph{IEEE transactions
  on neural networks and learning systems}, vol.~29, no.~4, pp. 1069--1081,
  2017.

\bibitem{20}
J.~Lin, L.~Wang, F.~Gao, S.~Shen, and F.~Zhang, ``Flying through a narrow gap
  using neural network: an end-to-end planning and control approach,''
  \emph{arXiv preprint arXiv:1903.09088}, 2019.

\bibitem{airsim}
S.~Shah, D.~Dey, C.~Lovett, and A.~Kapoor, ``Airsim: High-fidelity visual and
  physical simulation for autonomous vehicles,'' in \emph{Field and service
  robotics}.\hskip 1em plus 0.5em minus 0.4em\relax Springer, 2018, pp.
  621--635.

\bibitem{exploration-imit}
Y.~Guo, J.~Choi, M.~Moczulski, S.~Bengio, M.~Norouzi, and H.~Lee,
  ``Self-imitation learning via trajectory-conditioned policy for
  hard-exploration tasks,'' \emph{arXiv}, pp. arXiv--1907, 2019.

\bibitem{13}
D.~Shi, X.~Dai, X.~Zhang, and Q.~Quan, ``A practical performance evaluation
  method for electric multicopters,'' \emph{IEEE/ASME Transactions on
  Mechatronics}, vol.~22, no.~3, pp. 1337--1348, 2017.

\bibitem{13b}
P.~Lu and E.-J. van Kampen, ``Active fault-tolerant control for quadrotors
  subjected to a complete rotor failure,'' in \emph{2015 IEEE/RSJ International
  Conference on Intelligent Robots and Systems (IROS)}.\hskip 1em plus 0.5em
  minus 0.4em\relax IEEE, 2015, pp. 4698--4703.

\bibitem{2}
T.~Haarnoja, A.~Zhou, P.~Abbeel, and S.~Levine, ``Soft actor-critic: Off-policy
  maximum entropy deep reinforcement learning with a stochastic actor,''
  \emph{arXiv preprint arXiv:1801.01290}, 2018.

\bibitem{PPO}
J.~Schulman, F.~Wolski, P.~Dhariwal, A.~Radford, and O.~Klimov, ``Proximal
  policy optimization algorithms,'' \emph{arXiv preprint arXiv:1707.06347},
  2017.

\bibitem{DDPG}
T.~P. Lillicrap, J.~J. Hunt, A.~Pritzel, N.~Heess, T.~Erez, Y.~Tassa,
  D.~Silver, and D.~Wierstra, ``Continuous control with deep reinforcement
  learning,'' \emph{arXiv preprint arXiv:1509.02971}, 2015.

\bibitem{21}
D.~P. Kingma, T.~Salimans, and M.~Welling, ``Variational dropout and the local
  reparameterization trick,'' in \emph{Advances in neural information
  processing systems}, 2015, pp. 2575--2583.

\bibitem{DQL1}
H.~Hasselt, ``Double q-learning,'' \emph{Advances in neural information
  processing systems}, vol.~23, pp. 2613--2621, 2010.

\bibitem{DQL2}
H.~v. Hasselt, A.~Guez, and D.~Silver, ``Deep reinforcement learning with
  double q-learning,'' in \emph{Proceedings of the Thirtieth AAAI Conference on
  Artificial Intelligence}, 2016, pp. 2094--2100.

\bibitem{19}
Y.~Bengio, J.~Louradour, R.~Collobert, and J.~Weston, ``Curriculum learning,''
  in \emph{Proceedings of the 26th annual international conference on machine
  learning}, 2009, pp. 41--48.

\bibitem{FIREEVAQ}
J.~Sharma, P.-A. Andersen, O.-C. Granmo, and M.~Goodwin, ``Deep q-learning with
  q-matrix transfer learning for novel fire evacuation environment,''
  \emph{IEEE Transactions on Systems, Man, and Cybernetics: Systems}, 2020.

\bibitem{10}
P.~Christiano, Z.~Shah, I.~Mordatch, J.~Schneider, T.~Blackwell, J.~Tobin,
  P.~Abbeel, and W.~Zaremba, ``Transfer from simulation to real world through
  learning deep inverse dynamics model,'' \emph{arXiv preprint
  arXiv:1610.03518}, 2016.

\bibitem{11}
J.~Tan, T.~Zhang, E.~Coumans, A.~Iscen, Y.~Bai, D.~Hafner, S.~Bohez, and
  V.~Vanhoucke, ``Sim-to-real: Learning agile locomotion for quadruped
  robots,'' \emph{arXiv preprint arXiv:1804.10332}, 2018.

\bibitem{12}
C.~Finn, P.~Abbeel, and S.~Levine, ``Model-agnostic meta-learning for fast
  adaptation of deep networks,'' \emph{arXiv preprint arXiv:1703.03400}, 2017.

\bibitem{14}
O.~M. Andrychowicz, B.~Baker, M.~Chociej, R.~Jozefowicz, B.~McGrew,
  J.~Pachocki, A.~Petron, M.~Plappert, G.~Powell, A.~Ray \emph{et~al.},
  ``Learning dexterous in-hand manipulation,'' \emph{The International Journal
  of Robotics Research}, vol.~39, no.~1, pp. 3--20, 2020.

\bibitem{8}
J.~Hwangbo, J.~Lee, A.~Dosovitskiy, D.~Bellicoso, V.~Tsounis, V.~Koltun, and
  M.~Hutter, ``Learning agile and dynamic motor skills for legged robots,''
  \emph{Science Robotics}, vol.~4, no.~26, 2019.

\bibitem{15}
A.~Nichol, J.~Achiam, and J.~Schulman, ``On first-order meta-learning
  algorithms,'' \emph{arXiv preprint arXiv:1803.02999}, 2018.

\bibitem{ADR}
B.~Mehta, M.~Diaz, F.~Golemo, C.~J. Pal, and L.~Paull, ``Active domain
  randomization,'' in \emph{Conference on Robot Learning}.\hskip 1em plus 0.5em
  minus 0.4em\relax PMLR, 2020, pp. 1162--1176.

\end{thebibliography}

\end{document}